\documentclass[10pt,twocolumn,letterpaper]{article}

\usepackage{cvpr}              
\usepackage{multirow}
\usepackage{amsmath}
\usepackage{amssymb}
\usepackage{marvosym}

\usepackage{arydshln}
\usepackage{booktabs}
\usepackage{multirow}
\usepackage{enumitem}
\usepackage{pifont}
\usepackage{booktabs}
\usepackage{appendix}
\usepackage{stfloats}
\usepackage{graphicx}
\usepackage{wrapfig}

\usepackage[marginal]{footmisc}

\usepackage{float}

\definecolor{cvprblue}{rgb}{0.21,0.49,0.74}
\usepackage[pagebackref,breaklinks,colorlinks,allcolors=cvprblue]{hyperref}

\def\paperID{6963} 
\def\confName{CVPR}
\def\confYear{2025}

\title{~\includegraphics[height=20pt]{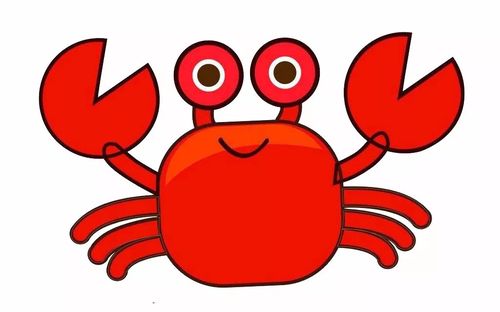}~Crab: A Unified Audio-Visual Scene Understanding Model \\ with Explicit Cooperation}

\author{
\textbf{Henghui Du}\textsuperscript{1,3,*},
\textbf{Guangyao Li}\textsuperscript{2},
\textbf{Chang Zhou}\textsuperscript{3},
\textbf{Chunjie Zhang}\textsuperscript{3},
\textbf{Alan Zhao}\textsuperscript{3},
\textbf{Di Hu}\textsuperscript{1,\Letter}
\\
\textsuperscript{1}Gaoling School of Artificial Intelligence, Renmin University of China, Beijing\\
\textsuperscript{2} Department of Computer Science and Technology, Tsinghua University, Beijing, China \\
\textsuperscript{3} AI Technology Center, Online Video Business Unit, Tencent PCG \\
\tt\small{\textsuperscript{1}\{cserdu,dihu\}@ruc.edu.cn, 
\textsuperscript{2}guangyaoli@tsinghua.edu.cn} \\
\tt\small{\textsuperscript{3}\{cserdu,chanzhou,Chunjie Zhang,Alan Zhao\}@tencent.com}
}

\begin{document}

\twocolumn[{
\renewcommand\twocolumn[1][]{#1}
\maketitle
\vspace{-2.25em}
\centering
\captionsetup{type=figure}\includegraphics[width=\linewidth]{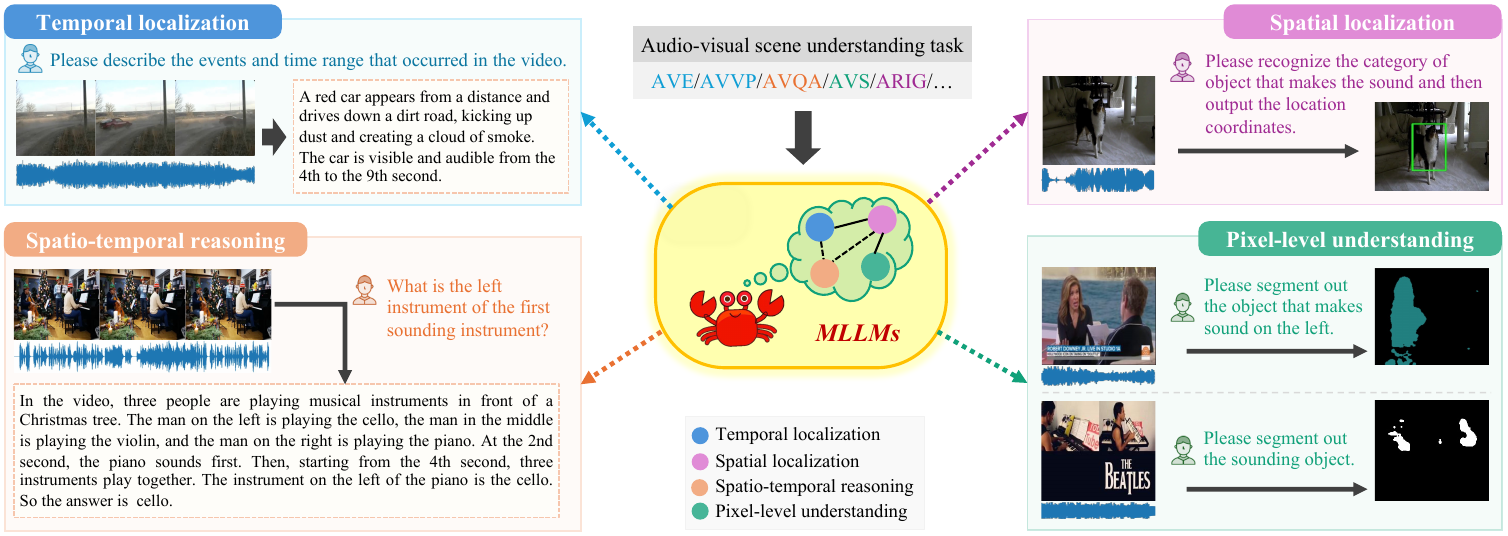}
\vspace{-2em}
\captionof{figure}{We present Crab, a unified audio-visual scene understanding model with explicit cooperation, which can complete various audio-visual tasks. It is trained on an instruction-tuning dataset with explicit reasoning process, which clarifies the cooperative relationship among tasks. Furthermore, to alleviate the interference caused by the learning process of complex audiovisual data and facilitate concrete cooperation, an interaction-aware LoRA structure is designed to enable the model focus on different aspects of data interaction.
}
\label{fig:teaser}
\vspace{1em}
}]

\footnote{
\textsuperscript{*}Intern at Tencent PCG, \textsuperscript{\Letter}Corresponding author.
}


\begin{abstract}
In recent years, numerous tasks have been proposed to encourage model to develop specified capability in understanding audio-visual scene, primarily categorized into temporal localization, spatial localization, spatio-temporal reasoning, and pixel-level understanding. Instead, human possesses a unified understanding ability for diversified tasks. Therefore, designing an audio-visual model with general capability to unify these tasks is of great value. However, simply joint training for all tasks can lead to interference due to the heterogeneity of audiovisual data and complex relationship among tasks. We argue that this problem can be solved through explicit cooperation among tasks. To achieve this goal, we propose a unified learning method which achieves \textbf{explicit inter-task cooperation} from both the perspectives of data and model thoroughly. Specifically, considering the labels of existing datasets are simple words, we carefully refine these datasets and construct an \textbf{A}udio-\textbf{V}isual \textbf{U}nified \textbf{I}nstruction-tuning dataset with \textbf{E}xplicit reasoning process (\textbf{AV-UIE}), which clarifies the cooperative relationship among tasks. Subsequently, to facilitate concrete cooperation in learning stage, an interaction-aware LoRA structure with multiple LoRA heads is designed to learn different aspects of audiovisual data interaction. By unifying the explicit cooperation across the data and model aspect, our method not only surpasses existing unified audio-visual model on multiple tasks, but also outperforms most specialized models for certain tasks. Furthermore, we also visualize the process of explicit cooperation and surprisingly find that each LoRA head has certain audio-visual understanding ability.
Code and dataset: 
\href{https://github.com/GeWu-Lab/Crab}
{https://github.com/GeWu-Lab/Crab}
\end{abstract}



\vspace{-0.5em}
\section{Introduction}
\label{sec:intro}



\textit{``Alone we can do so little; together we can do so much."} 
\begin{flushright} --- \emph{Helen Keller} \end{flushright}

Over the past few years, many works have explored the utilization of audio and visual modality to enhance general scene understanding, thereby deriving multiple tasks, primarily categorized into temporal localization, spatial localization, spatio-temporal reasoning, and pixel-level understanding. These tasks include audio-visual event localization (AVE)~\cite{tian2018audio}, audio-visual video parsing (AVVP)~\cite{tian2020unified}, audio referred image grounding (ARIG)~\cite{senocak2018learning}, audio-visual question answering (AVQA)~\cite{li2022learning}, audio-visual segmentation (AVS)~\cite{zhou2022audio} and reference audio-visual segmentation (Ref-AVS)~\cite{wang2024ref}, \emph{etc}. Specifically, AVE and AVVP tasks focus on temporal localization, where the model needs to accurately predict the events occurring in the video and their temporal boundaries. The purpose of ARIG task is to endow models with grounding capabilities, enabling them to locate the position of sounding objects. The AVQA task requires the model to integrate temporal and spatial information to answer questions. To complete AVS and Ref-AVS tasks, the model needs to output segmentation mask of target objects at pixel level. While significant advancements have been made in exploring these tasks, these works~\cite{hu2020discriminative,yu2022mm,sun2023learning,lin2023vision,li2024boosting,gao2024avsegformer,wang2024ref, hou2024toward, li2023multi} have mainly focused on completing specific task. By contrast, humans can generally possess a general perception and understanding for complex audio-visual scene, hence we also aspire to equip the model with similar capability. 

Unified general-purpose models can effectively transfer knowledge across tasks, and even perform unobserved tasks during training process. Early works primarily focused on unifying multiple language tasks, such as GPT2~\cite{radford2019language} and PaLM~\cite{chowdhery2023palm}. Subsequently, large language models (LLMs)~\cite{touvron2023llama,zeng2022glm,chiang2023vicuna} have demonstrated human-level performance across various language tasks. Then researchers further extend LLMs into multimodal domain. A line of works~\cite{li2023blip,ataallah2024minigpt4,chen2023minigpt} perform well on visual-language tasks. Recently, utilizing LLMs to complete multiple audio-visual tasks has become a goal pursued by the community~\cite{zhang2023video,cheng2024videollama,chowdhury2024meerkat}. They mainly accomplished this by naively curating instruction-tuning dataset and training jointly on multiple tasks, which is essentially an implicit approach to achieve inter-task cooperation. Considering the heterogeneity of audiovisual data and complex relationship among audio-visual tasks, it is difficult to more effectively achieve mutual cooperation among these tasks through such simple paradigm. For example, for two different types of tasks, temporal localization and spatial localization, instruction fine-tuning is effective when applied to one type of task individually, but directly training jointly on these two tasks could lead to interference.

We argue that this problem can be addressed through explicit cooperation among tasks from both the data and model perspectives, thoroughly. On the one hand, considering the labels of existing datasets are relatively simple words or phrases, 
they are insufficient for different tasks to establish relationship. 
Therefore, it is essential to augment existing datasets to explicitly clarify the cooperative relationship among tasks. As shown in Fig.~\ref{fig:teaser}, for spatio-temporal reasoning task, the label is no longer single words, but includes explicit reasoning process, such as the time when the first instrument makes a sound, the position of instruments, \emph{etc}. These information can help establish cooperation relationship among tasks. On the other hand, implicit methods may cause interference in the learning stage of heterogenous audiovisual data, hence the models needs to focus on different aspects of audiovisual data interaction, facilitating concrete cooperation in learning stage. As shown in Fig.~\ref{fig:teaser}, the model can focus on multiple interaction aspects such as temporal, spatial, and pixel-level, thereby facilitating concrete cooperation.

To achieve this goal, we propose a unified audio-visual scene understanding model with explicit cooperation. Specifically, we first refine existing multiple audio-visual task datasets and construct an \textbf{A}udio-\textbf{V}isual \textbf{U}nified \textbf{I}nstruction-tuning dataset with \textbf{E}xplicit reasoning process (\textbf{AV-UIE}), which clarifies the concrete cooperative relationship among tasks. The learning objective of task is not only to predict simple labels, but also includes explicit reasoning process, which contains information from multiple spatio-temporal dimensions. The model can rely on these information to establish relationship among tasks. Then, to alleviate the interference and facilitate concrete cooperation in learning stage, an interaction-aware LoRA structure with multiple LoRA heads is designed. Each head is responsible for learning certain audiovisual data interaction aspect. By unifying the explicit cooperation across data and model aspect, our method not only surpasses existing unified audiovisual models on multiple tasks, but also outperforms the specialized models for certain tasks. Furthermore, we also visualize the process of explicit cooperation and surprisingly find that each LoRA head has certain audio-visual understanding ability.

Our contributions can be summarized as follows:
\begin{itemize}
    \item Based on unification of audio-visual tasks, we propose an approach of explicit cooperation, which solves the problem of unified learning of complex audiovisual data.
    \item We construct AV-UIE, an audio-visual unified instruction-tuning dataset with explicit reasoning process, which clarifies the cooperation relationship among tasks.
    \item We design an interaction-aware LoRA structure to focus on different audiovisual data interaction aspects, thereby facilitating concrete cooperation in learning stage.
\end{itemize}

\section{Related Works}
\label{sec:related work}

\begin{figure*}[ht]
     \centering
     \includegraphics[width=0.9\textwidth]{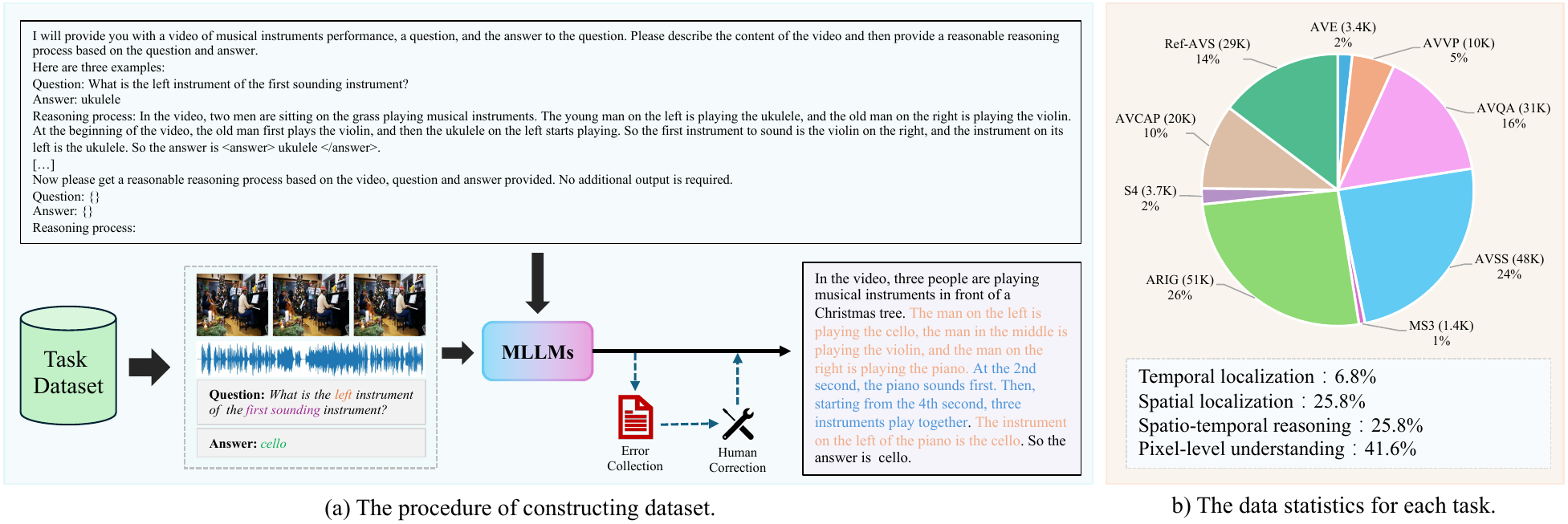}
     \vspace{-1em}
     \caption{
     Our proposed AV-UIE dataset. (a) explains the specific process of dataset construction, and (b) is the data analysis for all tasks.
     }
     \label{fig:dataset}
     \vspace{-1em}
\end{figure*}

\textbf{Unified Learning for Multi-task.} Pre-training language models on a large amount of unlabeled text corpora has become a standard practice, such as BERT~\cite{kenton2019bert}. To address certain task, the model needs to be fine-tuned on corresponding downstream task. Therefore, researchers have been interested in developing unified models to solve a wide range of language tasks~\cite{brown2020language,chowdhery2023palm}. Inspired by this success, many efforts~\cite{yang2022unitab,chen2022unified,chen2022pali,lu2022unified,lu2024unified} have been devoted to building unified models that can apply to vision tasks. In this work, we also use a unified interface to accomplish multiple tasks, but we explore more complex audio-visual scene understanding tasks.

\textbf{Audio-Visual Unified MLLMs.} Leveraging powerful reasoning capability of LLMs to accomplish multiple audio-visual tasks has become a recent trend~\cite{zhang2023video,cheng2024videollama,sun2024video,chowdhury2024meerkat}. VideoLLaMA 2~\cite{cheng2024videollama} integrated audio and video branch into model through joint training, thereby enriching the model's multimodal understanding capability. MEERKAT~\cite{chowdhury2024meerkat} builds a fine-grained large-scale audio-visual instruction-tuning dataset, endowing the model with grounding ability in time and space. While these works have made significant advancements, they simply construct instruction-tuning dataset and conduct joint training on audio-visual tasks. 
In contrast to these methods, we propose a unified learning method that achieves explicit cooperation.

\textbf{Multi-task LoRA.} LoRA~\cite{hu2021lora} is an approach that utilizes limited computational resources to efficiently finetune large language models. Researchers further explore the combination of multiple LoRAs to handle complex tasks. LoRAhub~\cite{huang2023lorahub} trains multiple LoRA adapters separately, and then selects appropriate combination of adapters during inference process to complete specific task. 
These works aim to use multiple LoRAs to address different downstream tasks. Unlike them, in a multi-task unified framework, to achieve the goal of explicit cooperation among audio-visual tasks, we design an interaction-aware LoRA structure to decouple the model's various capabilities.

\section{AV-UIE Dataset}
\label{sec:dataset}

\begin{figure*}[ht]
     \centering
     \includegraphics[width=0.9\textwidth]{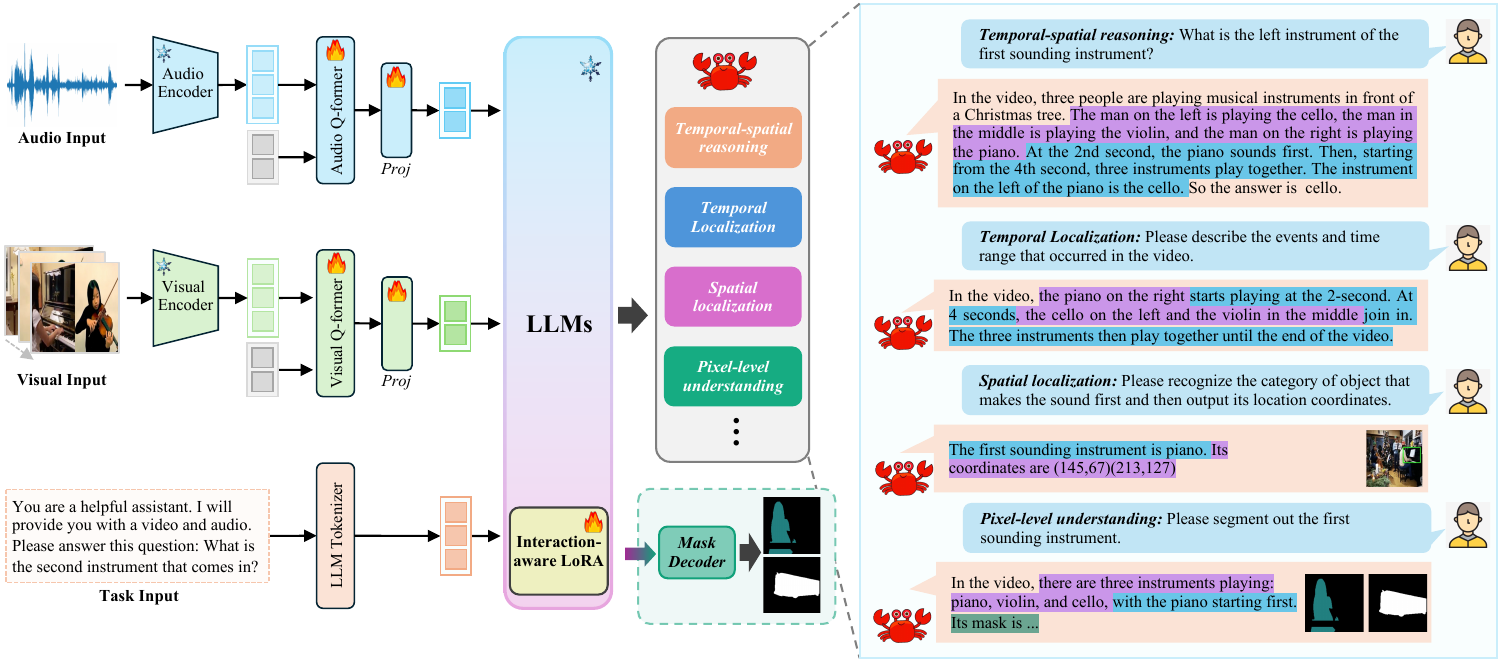}
     \vspace{-1em}
     \caption{
     The architecture of our unified audio-visual scene understanding model. It mainly consists of two parts: unified audio-visual interface, which consists of three multimodal branches, and a large language model with interaction-aware LoRA structure. The audio branch and visual branch process audio and video inputs respectively, while the segmentation branch is responsible for outputting the segmentation mask. The model is trained on our AV-UIE dataset, which clarifies the cooperation relationship among tasks, as marked by different colors on the right side of the figure. Content of same color in different tasks can help model establish cooperative relationship among tasks. Furthermore, to alleviate the interference caused by the learning process of complex audiovisual data, we design an interaction-aware LoRA structure to facilitate concrete cooperation.
     }
     \label{fig:framework}
     \vspace{-1.25em}
\end{figure*}


In order to clarify the explicit cooperative relationship and enable model to establish connections among tasks, we construct AV-UIE, an audio-visual unified instruction-tuning dataset with explicit reasoning process. In Section~\ref{construct}, we will introduce the construction process in detail, and in Section~\ref{analysis} we provide an analysis for our dataset.

\subsection{Dataset Construction}
\label{construct}
Our dataset is primarily an augmentation of existing audio-visual task datasets. 
Considering the raw data labels are single words or phrases, we transform these plain labels into an explicit reasoning process. Figure~\ref{fig:dataset}(a) illustrates a detailed construction procedure. Specifically, we first sample an instance from the dataset of each task, which includes audio, video, and simple labels. Then, using the in-context learning method, we prompt existing powerful MLLMs\footnote{We use Gemini 1.5 Pro to complete this process.} to output correct reasoning process based on provided information. In this process, we ensure that the final label after transformation remains consistent with original label. Finally, to ensure the quality of data, we filter out poor quality instances and use manual annotation for correction. For instance, in the example described in Figure~\ref{fig:dataset}(a), this video shows a scene of three people playing musical instruments. The question is ``What is the left instrument of the first sounding instrument?", and the corresponding answer is ``cello". Through our dataset construction process, a detailed reasoning process can be obtained, which contains rich information in terms of temporal and spatial, conducive to temporal and spatial localization tasks. 

Subsequently, following above construction procedure, we transform the MUSIC-AVQA~\cite{li2022learning}, AVE~\cite{tian2018audio} fully supervised, LLP~\cite{tian2020unified}, AVS-Bench~\cite{zhou2022audio} and Ref-AVS~\cite{wang2024ref} datasets. Specifically, for MUSIC-AVQA and AVE datasets, we directly transform the labels as described in construction process. Since the training data of LLP dataset only contains video-level event labels and lacks the temporal boundaries of event occurrence, We use the same MLLMs to annotate additional temporal information. The transformed LLP dataset can also benefit from data related to other temporal localization and spatio-temporal reasoning tasks, which contain precise temporal labels. For AVS-Bench and Ref-AVS datasets, we explicitly add information about the object making the sound. For spatial localization task, we collect data from the AVS-Bench dataset and obtain the bounding box of sounding object. Each box is represented by its top-left and bottom-right corners, i.e., [$x_{Left}, y_{Top}, x_{Right}, y_{Bottom}]$. To enhance the model's spatio-temporal ability, our dataset additionally includes the 
perception dataset proposed by VALOR~\cite{chen2023valor}. To this end, we can obtain an instruction-tuning dataset with explicit reasoning process, namely AV-UIE dataset. We provide prompt and examples in the \textit{supplementary materials}.

\vspace{-0.25em}
\subsection{Dataset Analysis}
\vspace{-0.25em}
\label{analysis}
Our dataset contains approximately $200K$ training samples with explicit reasoning process, as shown in Fig.~\ref{fig:dataset}(b). The data of  pixel-level understanding tasks account for the largest proportion, reaching 41.6\% (S4 $1.7\%$, MS3  $0.7\%$, AVSS $24.3\%$ and Ref-AVS $14.7\%$), while spatial localization (ARIG $25.8\%$) and spatio-temporal reasoning tasks (MUSIC-AVQA $15.7\%$ and AVCap $10.1\%$) account for 25.8$\%$ each. 
The number of temporal localization tasks is the smallest, accounting for 6.8\% (AVE $1.7\%$ and LLP $5.1\%$). 
Compared to previous instruction-tuning datasets~\cite{chowdhury2024meerkat,li2024groundinggpt,ren2024timechat}, our dataset covers a wider range of task types, including temporal, spatial localization, spatio-temporal reasoning, and pixel-level understanding tasks. To the best of our knowledge, it is the first dataset to cover all these audio-visual scene understanding tasks. Moreover, while the total amount of our data is relatively smaller for each task, thanks to the explicit reasoning process, we can still achieve superior results with a smaller amount of data.


\section{Method}
\label{sec:method}


In this section, we present a detailed introduction to our model. As depicted in Fig.~\ref{fig:framework}, our framework comprises two parts: 
\textit{{\romannumeral1}}) unified audio-visual interface, which consists of three multimodal branches, each responsible for processing audio, visual, and segmentation mask output, \textit{{\romannumeral2}}), a large language model with interaction-aware LoRA structure. 


\vspace{-0.5em}
\subsection{Unified Audio-Visual Interface}
\label{interface}

\textbf{Visual branch.} The visual branch comprises a visual encoder and a visual-language alignment module. We employ a pre-trained CLIP-ViT-L/14~\cite{radford2021learning} encoder $\mathcal{E}^{I}(\cdot)$ as our visual encoder. Given $T$ frames of video input $\mathcal{V}=\{I_{i} \in \mathbb{R}^{H \times W \times C}\}_{i=1}^{T}$ where $H, W, C$ represent the height, width and channels respectively, we extract the patch-level feature of each frame $f_v \in \mathbb{R}^{L_v \times D_v}$ where $L_v$ and $D_v$ denote the number of image tokens and feature dimension respectively. The visual embeddings of video can be denoted as $F_v = \{f_v^i\}_{i=1}^T$. Then we use a Q-Former~\cite{li2023blip} with $K_v$ learnable query tokens to compress the visual features into a fixed number of tokens. These visual tokens will be aligned to the textual semantic space through a two-layer MLP.

\noindent\textbf{Audio branch.} 
It consists of a pre-trained audio encoder and an audio-language alignment module. 
The audio encoder transforms the raw audio into an audio embedding. We use BEATs~\cite{chen2022beats} as audio encoder $\mathcal{E}^A(\cdot)$, which is trained to extract audio semantics information using iterative self-supervised learning. We select $T$ audio segments corresponding to the visual frames. For each audio segment $a_i$, we can obtain corresponding audio feature $f_a^i \in \mathbb{R}^{L_a \times D_a}$ where $L_a$ and $D_a$ are token numbers and feature dimension of audio feature. The audio embeddings can be denoted as $F_a = \{f_a^i\}_{i=1}^{T}$. Similarly, our audio-language alignment module also comprises a Q-Former followed by a two-layer MLP. Finally, we can get $K_a$ audio tokens aligned to the textual space where $K_a$ is the number of learnable query tokens. These tokens will be fed into the LLM.

\noindent\textbf{Segmentation branch.} Inspired by the success of SAM~\cite{kirillov2023segment} in image segmentation, recent works~\cite{ren2024pixellm,lai2024lisa} have combined the mask decoder in SAM with LLMs to complete segmentation task. Our segmentation branch mainly consists of a mask decoder $\mathcal{D}^M(\cdot)$ with a similar structure. To adapt the mask decoder to our model, we extend the original LLM vocabulary with multiple 
\texttt{<MASK>} tokens, i.e. $C_{seg} = \{\texttt{<MASK>}_i\}_{i=0}^{N}$ where $N$ is the number of tokens, which signify the segmentation output. When the model intends to generate a segmentation mask, the output would include these \texttt{<MASK>} tokens. We then extract the LLM last-layer embeddings corresponding to these \texttt{<MASK>} tokens. These embeddings serve as a prompt and will be sent to the mask decoder together with the visual embeddings $F_v$. Finally the mask decoder predicts the segmentation mask $\mathcal{M} \in \mathbb{R}^{C \times H \times W}$ where $C$ represents the number of categories, $H$ and $W$ represent height and width of mask image respectively. 



\vspace{-0.25em}
\subsection{Interaction-aware LoRA}
\label{lora}

LoRA~\cite{hu2021lora} is a highly efficient method for finetuning LLMs, enabling them to acquire specific capabilities. In a standard LoRA structure, there are two trainable pairs of rank decomposition matrices $A \in \mathbb{R}^{h \times r}$ and $B \in \mathbb{R}^{r \times h}$ where $h$ and $r$ is the hidden dimension of LLMs and LoRA rank. Rather than employing a single LoRA, multiple LoRA structure employs multiple independent and smaller LoRA, with each LoRA responsible for addressing a specific downstream task. 
Both structures have drawbacks in unifying multiple tasks. The former relies on a fully parameter-shared LoRA to implicitly learn the capabilities needed for all tasks, making it difficult to achieve effective task cooperation. The latter, with independent LoRAs for each task, hinders inter-task cooperation.

\begin{figure}[t]
    \centering
    \includegraphics[width=0.47\textwidth]{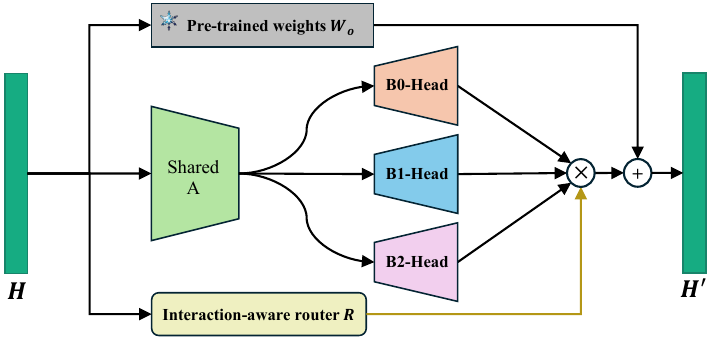}
     \caption{The architecture of interaction-aware LoRA structure. It comprises a shared matrix $A$ and multiple matrices $B$, also known as LoRA head. Matrix $A$ learns common multimodal representations, while each LoRA head is dedicated to learn certain audiovisual data interaction aspect. 
     }
     \label{fig:lora}
     \vspace{-2em}
\end{figure}

Inspired by HydraLoRA~\cite{tian2024hydralora}, in order to enable the model focus on different audiovisual data interaction aspects, we design an interaction-aware LoRA structure, a specialized fine-tuning architecture for multi-task. Fig.~\ref{fig:lora} provides a detailed explanation of this structure. It mainly consists of a shared parameter matrix $A$ and multiple independent matrices $B$, also known as LoRA heads. Different from HydraLoRA, the purpose of our interaction-aware LoRA structure is to alleviate interference of complex audiovisual data and facilitate concrete cooperation in learning stage. Under this structure, each head can focus on certain aspect of data interaction, with our expectations focusing on temporal, spatial, and pixel-level, \emph{etc}. Therefore, the number of LoRA heads is not related to the task numbers, but is determined by different types of data interaction.

\begin{table*}[tbp]
\belowrulesep=0pt
\aboverulesep=0pt
\begin{center}
\caption{The comparison results with other general models on all type of tasks. MS3 and AVSS are two subtasks of AVS-Bench. Seen is a subtask of the Ref-AVS test set. The X-InstructBLIP's performance on AVQA is zero-shot. \ding{51} indicates the model has ability to complete this type of task, but no evaluation is provided in their paper. \ding{55} indicates the model does not have the corresponding ability.}
\vspace{-1em}
\label{tab_uni_comp}
\scalebox{0.8}{
\begin{tabular}{c|c|cc|cc|c|cc|cc|cc}
\toprule
& \multicolumn{1}{c|}{\textbf{AVE}} & \multicolumn{2}{c|}{\textbf{AVVP}} & \multicolumn{2}{c|}{\textbf{ARIG}}  & \multicolumn{1}{c|}{\textbf{AVQA}} & \multicolumn{2}{c|}{\textbf{MS3(AVS)}} & \multicolumn{2}{c|}{\textbf{AVSS(AVS)}} & \multicolumn{2}{c}{\textbf{Seen(Ref-AVS)}} \\
\multirow{-2}{*}{\textbf{Method}} & \textbf{Acc} & \textbf{Segment-level} & \textbf{Event-level} & \textbf{cIoU} & \textbf{AUC} & \textbf{Acc} & \textbf{mIOU} & \textbf{F-score} & \textbf{mIOU} & \textbf{F-score}  & \textbf{mIOU}  & \textbf{F-score} \\
\hline
TimeChat~\cite{ren2024timechat} & \ding{51} & 51.28 & \ding{51} & \ding{55} & \ding{55} & \ding{55} & \ding{55} & \ding{55} & \ding{55} & \ding{55} & \ding{55} & \ding{55} \\
MEERKAT~\cite{chowdhury2024meerkat} & \ding{51} & \underline{54.96} & \ding{51} & \ding{51}& \ding{51} & \ding{51} & \ding{55} & \ding{55} & \ding{55} & \ding{55} & \ding{55} & \ding{55}\\
GroundingGPT~\cite{li2024groundinggpt} & \ding{51} & \ding{51} & \ding{51} & \textbf{44.02} & \textbf{0.45} & \ding{51} & \ding{55} & \ding{55} & \ding{55} & \ding{55} & \ding{55} & \ding{55} \\
X-InstructBLIP~\cite{panagopoulou2023x} & \ding{55} & \ding{55} & \ding{55} & \ding{55} & \ding{55} & 44.50 & \ding{55} & \ding{55} & \ding{55} & \ding{55} & \ding{55} & \ding{55} \\
VALOR~\cite{chen2023valor} & \ding{55} & \ding{55} & \ding{55} & \ding{55} & \ding{55} & \underline{78.90} & \ding{55} & \ding{55} & \ding{55} & \ding{55} & \ding{55} & \ding{55} \\
AnyRef~\cite{he2024multi} & \ding{55}  & \ding{55} & \ding{55} & \ding{55} & \ding{55} & \ding{55} & \underline{55.6} & \textbf{66.30} & \ding{51} & \ding{51} & \ding{51} & \ding{51} \\
\hline
\textbf{Crab(Ours)} & \textbf{80.15} & \textbf{59.00} & \textbf{54.44} & \underline{41.78} & \underline{0.42} & \textbf{78.94} & \textbf{58.21} & \underline{66.24} & \textbf{26.52} & \textbf{32.10} & \textbf{40.54} & \textbf{0.58} \\
\bottomrule
\end{tabular}
}
\end{center}
\vspace{-2em}
\end{table*}


Specifically, $T$ frames of video input and corresponding audio segments first pass through the audio and visual branch to obtain corresponding representations, then they are concatenated and fed into the LLMs together with the task instruction as the initial hidden states $H_0$:
\begin{gather}
H_v = \text{MLP}_v(\text{Q-Former}_v(\mathcal{E}^{I}(F_v))) \in \mathbb{R}^{T K_v \times h}, \\
H_a = \text{MLP}_a(\text{Q-Former}_a(\mathcal{E}^{A}(F_a))) \in \mathbb{R}^{T K_a \times h},\\
H_0 = g([H_v \circ H_a \circ H_t ]) \in \mathbb{R}^{T (K_v + K_v) \times h},
\end{gather}
where $\circ$ denotes concatenation operation, $H_t$ is the text embedding. Then, $H_0$ will forward through each transformer block in the LLMs. In each block, similar to the common use of LoRA, our interaction-aware LoRA structure is added as a bypass branch in all linear layers. For each linear layer, there is a pre-trained and frozen weight $W_o$. The interaction-aware LoRA structure learns the change of $W_o$, denoted as $\Delta W$. Given the input hidden embedding $H \in \mathbb{R}^{T (K_v + K_v) \times h}$, we use $n$ heads $\{B_i\}_{i=1}^{n}$ to learn $\Delta W$, each head is responsible for learning a certain audiovisual data interaction aspect. In order to enable the model to determine how to effectively focus on these aspects, we additionally add an interaction-aware router structure $R$, which consists of a dense layer with trainable transformation matrix $W_r \in \mathbb{R}^{T (K_v + K_v) \times n}$ and a $softmax(\cdot)$ function to calculate the route scores $S = \{s_i\}_{i=1}^{n}$ of different interaction aspects. The output of this bypass branch is obtained by the weighted sum of all heads. 
This is formulated as:
\begin{gather}
    S = softmax(W_r H) \in \mathbb{R}^{T (K_v + K_v) \times n},\\
    \Delta W = \sum_{i=1}^n s_i B_i A H \in \mathbb{R}^{T (K_v + K_v) \times h},\\
    H' = W_o H + \Delta W,
\end{gather}
where $H'$ is the output of this linear layer, $A$ is a shared matrix to learn the common multimodal representation.

\vspace{-0.5em}
\subsection{Training Objective}
\label{objective}
Our overall training objective consists of two parts: the auto-regressive cross-entropy loss $\mathcal{L}_{txt}$ for text generation of all tasks, and the auxiliary loss required for AVS and Ref-AVS tasks $\mathcal{L}_{seg}$. Specifically, we use binary cross-entropy loss $\mathcal{L}_{bce}$ and $\mathcal{L}_{dice}$ for binary mask prediction and cross-entropy loss $\mathcal{L}_{ce}$ for the AVSS task. The final training loss can be denoted as follows:
\begin{gather}
\mathcal{L}_{seg} = \lambda_{bce} \cdot \mathcal{L}_{bce} + \lambda_{dice} \cdot \mathcal{L}_{dice} + \lambda_{ce} \cdot \mathcal{L}_{ce},\\
\mathcal{L} = \lambda_{txt} \cdot \mathcal{L}_{txt} + \lambda_{seg} \cdot \mathcal{L}_{seg}.
\end{gather}
Here, $\lambda_{txt}$, $\lambda_{seg}$, $\lambda_{bce}$, $\lambda_{dice}$, and $\lambda_{ce}$ are hyper-parameters.

\section{Experiments}
\label{sec:exp}
\subsection{Implement Details}
Each video is sampled uniformly with 10 frames, and the size of each frame is $224 \times 224$. We extract the last layer patch-level representations for each frame. Following BEATs~\cite{chen2022beats}, we convert the sample rate of each raw waveform to $16KHz$ and extract 128-dimensional Mel-filter bank features with 25ms Povey window that shifts every 10ms as the acoustic feature. For the mask decoder, we use two \texttt{<MASK>} token groups corresponding to two scales of visual features from visual encoder, and we utilize the visual features from the 14th and second-to-last layers. Each group has three tokens. The number of query tokens in both audio Q-Former and visual Q-Former is 32. We use the LLaMA-2-7b-Chat model as our base model. The interaction-aware LoRA structure is employed in all linear layers with a rank of 8. The hyper-parameters $\lambda_{txt}$, $\lambda_{seg}$, $\lambda_{bce}$, $\lambda_{dice}$, and $\lambda_{ce}$ are set to $1.0, 0.5, 1.0, 0.5, 1.0$ respectively. 
More details in the \textit{supplementary materials}.

\subsection{Training Procedure}
We consider a two-stage training paradigm, including pre-training for feature alignment and instruction-tuning for audio-visual tasks.

\textbf{Stage 1: Pre-training for feature alignment.} The purpose of pre-training is to train three multimodal branches in the unified audio-visual interface. We use pre-training data from Video-LLaVA~\cite{lin2023video} to train visual Q-Former and MLP layer, the audio-caption pairs data from AudioCaps~\cite{kim2019audiocaps} to train audio Q-Former and MLP layer, and image segmentation data from LVIS~\cite{gupta2019lvis} to train mask decoder. During training process, the weights of visual encoder and audio encoder are frozen. We first pretrain the visual branch and audio branch, and then train the segmentation branch with frozen visual branch weight. We train for three epochs with a global batch size of 256, using the AdamW optimizer with a cosine learning rate schedule. The initial learning rate is set to $1e-4$ with a warmup ratio of 0.03.

\textbf{Stage 2: Instruction-tuning for audio-visual Tasks.} In the second stage, we train all downstream tasks on our AV-UIE dataset. The trainable parameters include three multimodal branches (excluding the audio encoder and visual encoder) and interaction-aware LoRA. The data in each batch is a random combination from different tasks. We train for five epochs with a global batch size of 512. The remaining settings are the same as the first stage. We provide additional hyper-parameters in the \textit{supplementary materials}.

\begin{table}[t]
\belowrulesep=0pt
\aboverulesep=0pt
\centering
\caption{The comparison results with specialized models on temporal localization task.}
\vspace{-1em}
\label{tab_ave_avvp}
\setlength{\tabcolsep}{2.75mm}{
\scalebox{0.8}{
\begin{tabular}{c|c|cc}
\bottomrule
\multirow{2}{*}{\textbf{Method}} & \textbf{AVE task} & \multicolumn{2}{c}{\textbf{AVVP task}} \\
& \textbf{Acc} & \textbf{Segment-level} & \textbf{Event-level} \\
\hline
AVT~\cite{lin2020audiovisual} & 75.80 & - & - \\
PSP~\cite{zhou2021positive} & \underline{77.80} & - & - \\
MM-Pyramid~\cite{yu2022mm} & \underline{77.80} & 59.20 & 53.04 \\
CMBS~\cite{xia2022cross} & - & 55.00 & 48.48 \\
MPN~\cite{yu2021mpn} & 77.60 & - & - \\
DHHN~\cite{jiang2022dhhn} & - & \textbf{60.32} & \textbf{55.06} \\
\hline
\textbf{Crab(Ours)} & \textbf{80.15} & 59.00 & \underline{54.44} \\
\toprule
\end{tabular}
}
}
\vspace{-1em}
\end{table}

\begin{table}[htbp]
\belowrulesep=0pt
\aboverulesep=0pt
\begin{center}
\caption{The comparison results with specialized models on MUSIC-AVQA test set.}
\vspace{-1em}
\label{tab_avqa}
\setlength{\tabcolsep}{3.2mm}{
\scalebox{0.8}{
\begin{tabular}{c|c|c|c|c}
\toprule
\textbf{Method} & \textbf{Audio}  & \textbf{Visual} & \textbf{Audio-Visual}  & \textbf{Avg} \\
\hline
ST-AVQA~\cite{li2022learning} & 73.87 & 74.40 &  69.53 & 71.59  \\
COCA~\cite{lao2023coca} &  75.42 & 75.23  & 69.96 & 72.33  \\
PSTP-Net~\cite{li2023progressive} & 70.91 & 77.26 & 72.57 & 73.52 \\
LAVISH~\cite{lin2023vision} & 75.97 & 80.22 & 71.26 & 74.46 \\
TSPM~\cite{li2024boosting}  & \textbf{76.91} & \underline{83.61} & \underline{73.51} & \underline{76.79} \\
\hline
\textbf{Crab(Ours)} & \underline{76.58} & \textbf{90.73} &  \textbf{74.13} & \textbf{78.94} \\ 
\bottomrule
\end{tabular}
}
}
\end{center}
\vspace{-2em}
\end{table}

\subsection{Quantitative Results and Analysis}
To demonstrate effectiveness of our method, we first compare with other general models across all tasks, and then compare with task-specialized models on each task. For temporal localization task, we evaluate on AVE~\cite{tian2018audio} fully supervised and LLP~\cite{tian2020unified} test set. For spatio-temporal reasoning task, we evaluate on MUSIC-AVQA test set. For spatial localization task, we adapt AVS-Bench~\cite{zhou2022audio} S4 test set to construct an audio referred image grounding test set. 
For pixel-level understanding task, we compare on AVS~\cite{zhou2022audio} and Ref-AVS~\cite{wang2024ref} test set respectively. We present main metrics for AVQA, AVS, and Ref-AVS tasks. Top-2 results are highlighted. 
In addition to zero-shot results mentioned in Tab.~\ref{tab_uni_comp}, all baseline models include training data for corresponding tasks to ensure fair assessment.
More metric results in the \textit{supplementary materials}.

\textbf{Comparison with other general models.} For general models, we keep consistent with foundational LLMs used by official, including LLaMA2-7B, Vicuna-v1.5. Tab.~\ref{tab_uni_comp} shows the comparison results of our model with other general models on all types of tasks.
As shown, our model outperforms other general-purpose models by handling a broader range of audio-visual scene understanding tasks and delivering superior performance across multiple tasks.
Specifically, we achieve a significant performance improvements on AVVP tasks (59.00\% \textit{vs.} 54.96\%). Compared to VALOR~\cite{chen2023valor}, which was trained on VALOR-1M dataset, our model achieves comparable results (78.94\% \textit{vs.} 78.90\%) while utilizing a smaller dataset. Since GroundingGPT~\cite{li2024groundinggpt} focuses on the general ability of fine-grained grounding task, its performance on ARIG task is slightly better than our model. On pixel-level understanding tasks, compared to AnyRef~\cite{he2024multi}, we can accomplish semantic segmentation task (AVSS) and reference segmentation tasks (Ref-AVS). And we also achieve superior result on MS3 task for mIOU metric and comparable result for F-score metric.

\textbf{Temporal localization task.} Tab.~\ref{tab_ave_avvp} shows the event localization accuracy, segment-level and event-level F1-score on temporal localization task. It is evident that our model achieves superior performance compared to specialized methods. Compared to MM-Pyramid~\cite{yu2022mm}, we achieves a performance improvement of 2.35\% (80.15\% \textit{vs.} 77.80\%). For AVVP task, our model achieves the comparable performance (59.00\% and 54.44\%). As discussed in Section~\ref{sec:dataset}, while temporal localization task accounts for the smallest proportion in our AV-UIE dataset, explicit inter-task cooperation enables this task benefit from other related audio-visual tasks, leading to improved performance.




\begin{table}[t]
\belowrulesep=0pt
\aboverulesep=0pt
\begin{center}
\caption{The comparison results with specialized models on spatial localization task.}
\vspace{-1em}
\label{tab_arig}
\setlength{\tabcolsep}{2.75mm}{
\scalebox{0.8}{
\begin{tabular}{c|c|c|c|c}
\toprule
\textbf{Method} & LVS~\cite{chen2021localizing} & EZ-VSL~\cite{mo2022localizing} & FNAC~\cite{sun2023learning} & \textbf{Crab(Ours)} \\
\hline
\textbf{cIoU} & 23.69 & 26.43 & \underline{27.15} & \textbf{41.78} \\
\textbf{AUC} & 0.25 & 0.29 & \underline{0.31} & \textbf{0.42} \\
\bottomrule
\end{tabular}
}
}
\end{center}
\vspace{-2em}
\end{table}

\begin{table}[t]
\belowrulesep=0pt
\aboverulesep=0pt
\begin{center}
\caption{The comparison results with specialized models on AVS-Bench and Ref-AVS. S4, MS3 and AVSS are the subtasks of AVS-Bench. Seen, Unseen and Null are the subtasks of Ref-AVS.}
\vspace{-1em}
\label{tab_avs_ref_avs}
\scalebox{0.7}{
\begin{tabular}{c|c|c|c|c|c|c}
\toprule
\textbf{Method} & 
\textbf{Backbone} &         
\textbf{MS3} &         
\textbf{AVSS} &
\textbf{Seen} &
\textbf{Unseen} & 
\textbf{Null($\downarrow$)} \\
\hline
AVSBench~\cite{zhou2022audio} & ResNet-50  & 54.00 & -  & 0.51 & 0.55 & 0.21 \\
TPAVI~\cite{zhou2022audio} & PVT-v2 & 54.00 & \textbf{29.80}  & -  & -  & - \\
AVSegFormer~\cite{gao2024avsegformer} & PVT-v2 & \textbf{58.40} & 24.90 & 33.47 & 36.05 & 0.17 \\
GAVS~\cite{wang2024prompting} & PVT-v2 & - & - & 28.93 & 29.82 & 0.19  \\
EEMC~\cite{wang2024ref} & PVT-v2 & - & - & \underline{34.20} & \textbf{49.54} & \textbf{0.01} \\ 
\hline
\textbf{Crab(Ours)} & \textbf{ViT/L-14} & \underline{58.21} & \underline{26.59} & \textbf{40.54} & \underline{45.55} & \textbf{0.01} \\ 
\toprule
\end{tabular}
}
\end{center}
\vspace{-1.5em}
\end{table}

\begin{table}[!t]
\belowrulesep=0pt
\aboverulesep=0pt
\begin{center}

\caption{The ablation results of explicit reasoning process (denoted as ERP) and interaction-aware LoRA (denoted as IA-LoRA).}
\label{tab_ablation}
\vspace{-1em}
\scalebox{0.775}{
\begin{tabular}{c|c|c|cc|cc}
\toprule
\multirow{2}{*}{\textbf{Method}} & \multicolumn{1}{c|}{\textbf{AVQA}} & \multicolumn{1}{c|}{\textbf{AVE}} & \multicolumn{2}{c|}{\textbf{AVVP}} & \multicolumn{2}{c}{\textbf{ARIG}} \\
&\textbf{Avg} & \textbf{Acc} & \textbf{Segment} & \textbf{Event} & \textbf{cIoU} & \textbf{AUC} \\
\hline
\textit{w/o.} ERP  & 76.05 & 78.62 & 52.01 & 51.36 & 40.92 & 0.41 \\
\textit{w/o.} IA-LoRA & 76.92 & 79.93 &  53.54 & 53.15 &  40.22 & 0.40  \\
\textbf{Crab(Ours)} & \textbf{78.94} & \textbf{80.15} & \textbf{59.00} & \textbf{54.44} & \textbf{41.78} & \textbf{0.42} \\ 

\bottomrule
\end{tabular}
}

\end{center}
\vspace{-2.5em}
\end{table}

\textbf{Spatial localization and pixel-level understanding task.} Tab.~\ref{tab_arig} shows the experiment results on ARIG task. From the table, it can be seen that our model achieves superior results compared with specialized models. To validate the model's pixel-level understanding capability, we present the experimental results on two representative tasks, AVS and Ref-AVS. In order to maintain a unified audio-visual interface for integration with other tasks, we chose not to use the widely adopted PVT-V2~\cite{wang2022pvt} model tailoredly designed for segmentation tasks as the visual backbone. Instead, we utilized ViT/L-14 to obtain visual features as input for mask decoder. As illustrated in Tab.~\ref{tab_avs_ref_avs}, our model achieves comparable performance on the MS3, AVSS subtasks and Unseen test set. It performs best on the Seen and Null test set. 

\textbf{Spatio-temporal reasoning task.} Tab.~\ref{tab_avqa} shows the experimental results on MUSIC-AVQA~\cite{li2022learning} test set. It can be observed that our method outperforms all specialized models. Specifically, compared to recent TSPM~\cite{li2024boosting}, our model achieves significant overall performance improvements of 2.15\% (78.94\% \textit{vs.} 76.79\%). On visual subtask, our method achieves remarkable improvements of 7.12\% (90.73\% \textit{vs.} 83.61\%). For complex audio-visual question type, our model obtains the best overall performance. And the performance in audio subtask is also comparable.






\subsection{Ablation Results}
To verify the effectiveness of explicit reasoning process and interaction-aware LoRA structure, we conduct two ablation experiments, including using only simple labels from original dataset and using a standard LoRA. Tab.~\ref{tab_ablation} presents the corresponding experiment results. When model does not output explicit reasoning process, the performance on AVQA, AVE, and AVVP tasks decreases significantly, while for the ARIG, AVS, and Ref-AVS tasks, the impact is minimal. 
This is primarily because the former task demands simultaneous comprehension of complex video and audio information, which heavily depends on the reasoning process. In contrast, the latter primarily relies on the model’s pixel-level understanding of visual content, where reasoning is not the main factor limiting performance. Furthermore, our interaction-aware LoRA structure consistently outperforms standard LoRA across all tasks.
More results for other tasks in the \textit{supplementary materials}.

\begin{table}[!t]
\belowrulesep=0pt
\aboverulesep=0pt
\begin{center}
\caption{The LoRA head drop results on interaction-aware LoRA. \textit{drop head-1} means the weight of \textit{head-1} is set to zero during inference process. And so on for the others.}
\vspace{-1em}
\label{tab_ablation_lora}
\setlength{\tabcolsep}{5.2mm}{
\scalebox{0.8}{
\begin{tabular}{c|c|cc|c}
\toprule
\multirow{2}{*}{\textbf{Method}} & \multicolumn{1}{c|}{\textbf{AVE}} & \multicolumn{2}{c|}{\textbf{ARIG}} & \multicolumn{1}{c}{\textbf{AVQA}} \\
& \textbf{Acc} & \textbf{cIoU} & \textbf{AUC} & \textbf{Acc} \\
\hline
\textit{drop head-1} & 75.07 & 36.15 & 0.36 & 76.33 \\
\textit{drop head-2} & 79.25 & 29.16 & 0.29 & 76.22 \\
\textit{drop head-3} & 79.62 & 38.65 & 0.39 & 77.01\\
\textbf{no drop} & \textbf{80.15} & \textbf{41.78} & \textbf{0.42} & \textbf{78.94} \\

\toprule
\end{tabular}
}
}
\end{center}
\vspace{-2em}
\end{table}

\subsection{Interaction-aware LoRA Analysis}
\begin{figure}[t]
    \vspace{-1em}
  \centering
  \begin{subfigure}[b]{0.23\textwidth}
    \includegraphics[width=\textwidth]{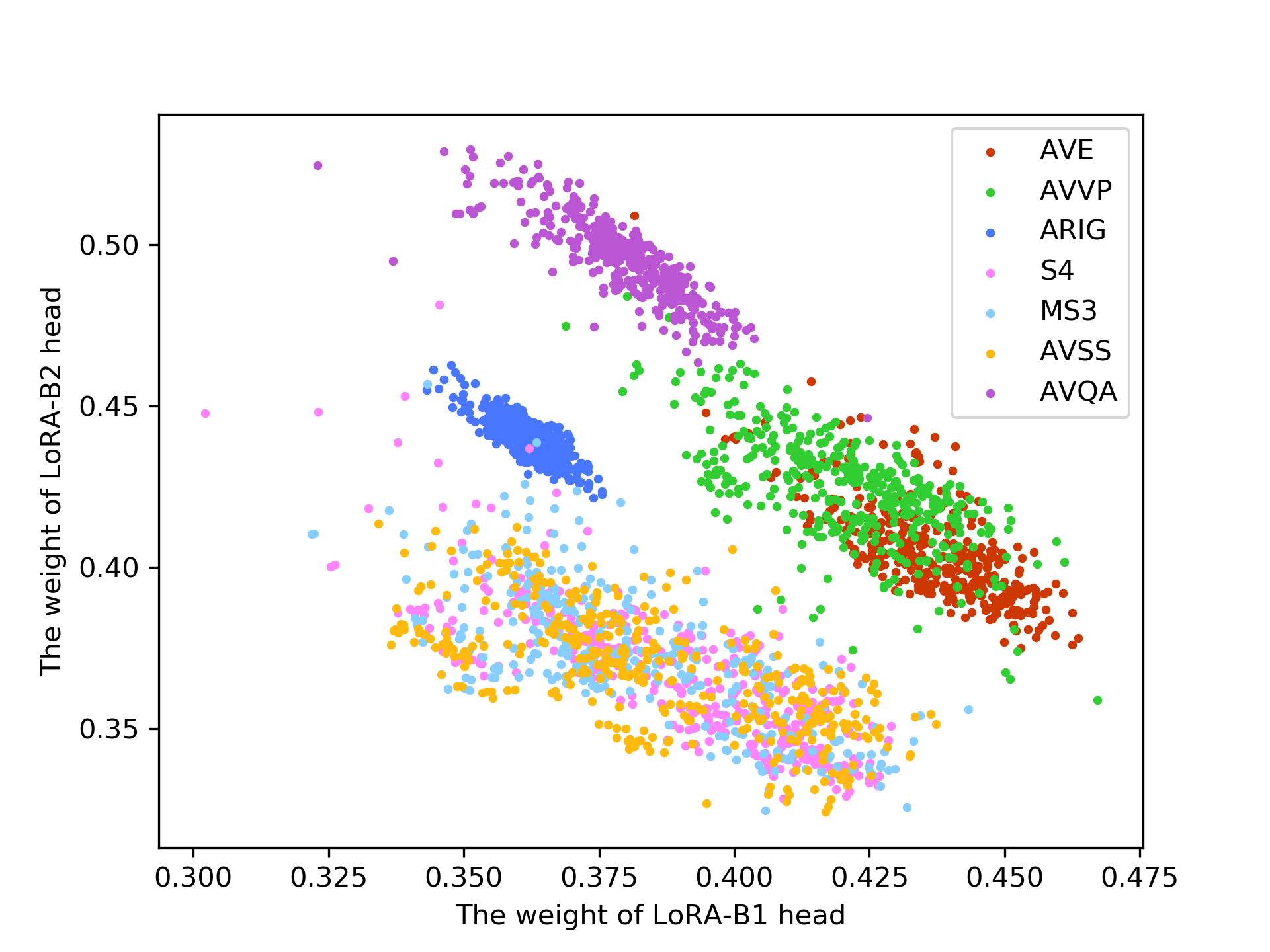}
    \caption{The router weight of \textit{LoRA-B1} and \textit{LoRA-B2} head.}
    \label{fig:route-01}
  \end{subfigure}
  \hfill 
  \begin{subfigure}[b]{0.23\textwidth}
    \includegraphics[width=\textwidth]{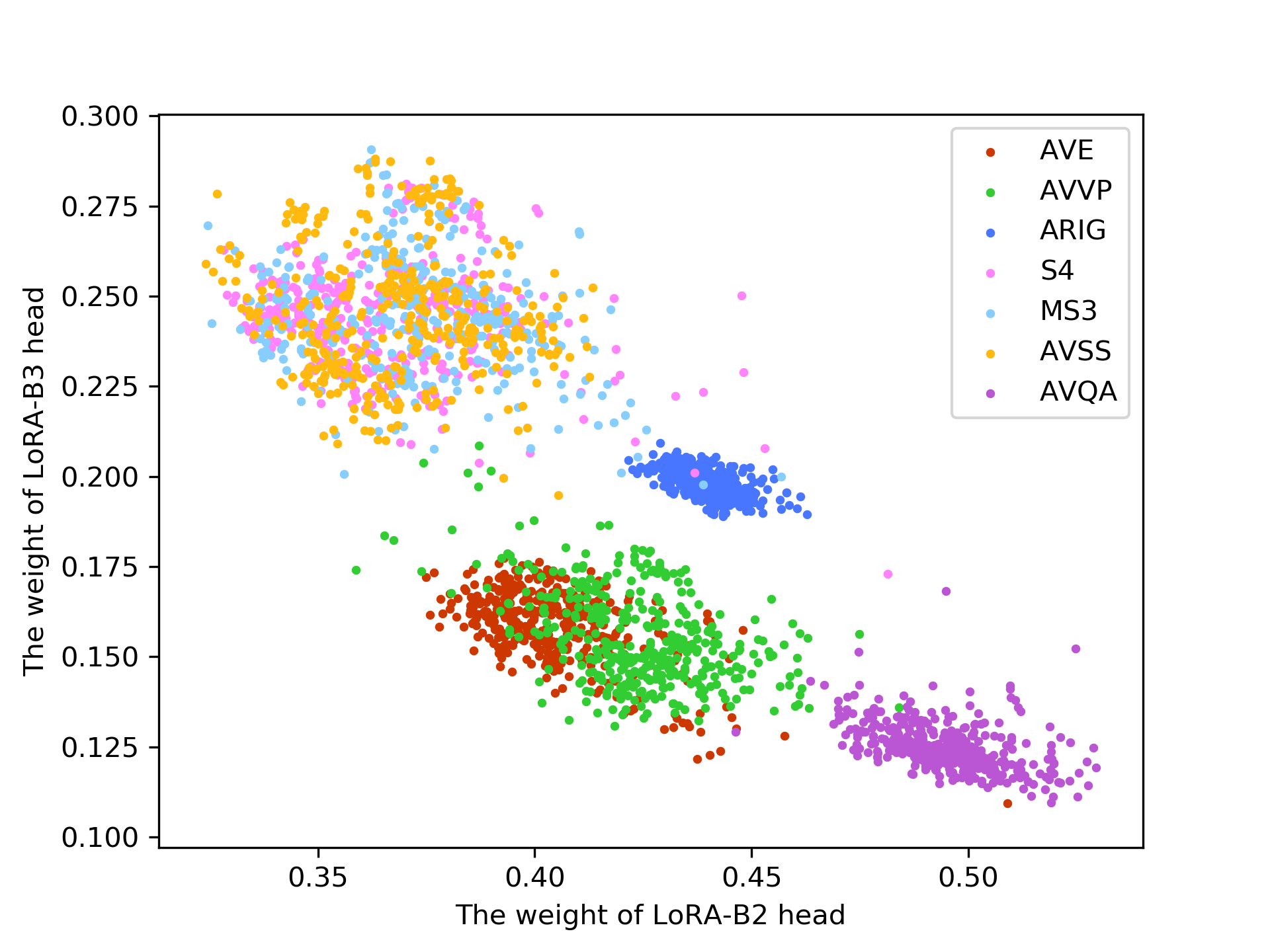}
    \caption{The router weight of \textit{LoRA-B2} and \textit{LoRA-B3} head.}
    \label{fig:route-12}
  \end{subfigure}
  \vspace{-0.5em}
  \caption{We visualize the router weights of three LoRA heads on different tasks. Figure (a) compares \textit{head-B1} and \textit{head-B2}, while figure (b) compares \textit{head-B2} and \textit{head-B3}. Different colors distinguish between tasks. The larger the router weight value, the greater the task's dependence on that LoRA head, indicating this LoRA head has a stronger ability to solve this type of task.}
  \label{fig:route}
  \vspace{-1.5em}
\end{figure}

\textbf{Visualized results.} 
To understand how tasks cooperate, we visualize the dependency of different tasks on LoRA heads during the inference process. 
As shown in Fig.~\ref{fig:route}, two key observations emerge: (1) Tasks of the same type exhibit similar dependencies on various LoRA heads. 
For instance, the temporal localization tasks like AVE and AVVP, the pixel-level understanding tasks like S4, MS3, and AVSS, 
tend to cluster together, while other task types, including AVQA and ARIG, form distinct clusters. 
It should be noted that this phenomenon is not caused by the homogeneity of data from tasks of the same type. The most intuitive example is our ARIG dataset, which is derived from AVS-Bench data, but these tasks do not form the same cluster. Moreover, the data from S4 and MS3 are not homogenous. 
However, they still cluster together because they belong to the same type of task and require similar skills to complete.
2) The varying dependence of different types of tasks on the three LoRA heads surprisingly suggests that these heads possess distinct capabilities. 
It can be found that the AVE and AVVP tasks rely significantly more on \textit{head-B1} and \textit{head-B2} than on \textit{head-B3}. 
The ARIG task is most dependent on \textit{head-B2}, followed by \textit{head-B1}, and finally \textit{head-B3}. Therefore, we can infer that \textit{head-B1} has stronger temporal localization capability, \textit{head-B2} has stronger spatial localization capability, and \textit{head-B3} is responsible for learning pixel-level understanding capability. Similarly, since the AVQA task requires stronger spatio-temporal reasoning capability, it also mainly relies on \textit{head-B1} and \textit{head-B2}, and less on \textit{head-B3}. 
Notably, the AVS task primarily relies on \textit{head-B1} and \textit{head-B2}, with a lesser dependence on \textit{head-B3}. This distribution may also contribute to the model's performance on the AVS task.
Despite this, compared to other tasks, the AVS task still has the greater dependence on \textit{head-B3}, reflecting that \textit{head-B3} does have pixel-level understanding capability, although it is not very strong.

\textbf{Drop LoRA head results.} Through above analysis, each LoRA head indeed possesses different capabilities. Furthermore, during the inference process, we set the weights of three LoRA heads to zero respectively, also known as ``\textit{drop head}". We find that setting weight directly to zero would affect the model's behavior to some extent. Therefore, we only report the comparison results under the model's normal behavior. The Tab.~\ref{tab_ablation_lora} presents corresponding results. From the table, we can observe that for temporal localization, spatial localization, and spatio-temporal reasoning tasks, dropping any head will result in a performance decline. Specifically, for AVE task, when we drop the head with temporal localization capability, namely \textit{head-1}, the performance is the worst. For the ARIG task, the performance is the worst when we drop \textit{head-2}. As for the AVQA task, the impact of these two heads on performance is relatively similar.

\vspace{-0.25em}
\section{Conclusion}
\vspace{-0.25em}
\label{sec:conclusion}
In this paper, we propose a unified audio-visual scene understanding model with explicit cooperation. To achieve this goal, we construct AV-UIE dataset, which clarifies the explicit cooperative relationship among tasks. Then we design an interaction-aware LoRA structure with multiple heads, each head is responsible for learning certain audiovisual data interaction aspect. Furthermore, we are surprised to discover that each head possesses certain capability through the visualized experimental results.

In future work, we plan to further enhance the model's pixel-level understanding capability. Additionally, while we have established connections among tasks through explicit reasoning process, its accuracy is not guaranteed, which can impact model's performance. Therefore, another possible major research direction is to enhance the reliability of reasoning process, thereby further improving the overall performance of the model. 

\noindent\textbf{Acknowledgement} The project was supported by National Natural Science Foundation of China (NO.62106272) and sponsored by CCF-Zhipu.AI Large Model Innovation Fund.

{
    \small
    \bibliographystyle{ieeenat_fullname}
    \bibliography{main}

\begin{thebibliography}{65}
\providecommand{\natexlab}[1]{#1}
\providecommand{\url}[1]{\texttt{#1}}
\expandafter\ifx\csname urlstyle\endcsname\relax
  \providecommand{\doi}[1]{doi: #1}\else
  \providecommand{\doi}{doi: \begingroup \urlstyle{rm}\Url}\fi

\bibitem[Ataallah et~al.(2024)Ataallah, Shen, Abdelrahman, Sleiman, Zhu, Ding, and Elhoseiny]{ataallah2024minigpt4}
Kirolos Ataallah, Xiaoqian Shen, Eslam Abdelrahman, Essam Sleiman, Deyao Zhu, Jian Ding, and Mohamed Elhoseiny.
\newblock Minigpt4-video: Advancing multimodal llms for video understanding with interleaved visual-textual tokens.
\newblock \emph{arXiv preprint arXiv:2404.03413}, 2024.

\bibitem[Brown(2020)]{brown2020language}
Tom~B Brown.
\newblock Language models are few-shot learners.
\newblock \emph{arXiv preprint arXiv:2005.14165}, 2020.

\bibitem[Chen et~al.(2021)Chen, Xie, Afouras, Nagrani, Vedaldi, and Zisserman]{chen2021localizing}
Honglie Chen, Weidi Xie, Triantafyllos Afouras, Arsha Nagrani, Andrea Vedaldi, and Andrew Zisserman.
\newblock Localizing visual sounds the hard way.
\newblock In \emph{Proceedings of the IEEE/CVF conference on computer vision and pattern recognition}, pages 16867--16876, 2021.

\bibitem[Chen et~al.(2023{\natexlab{a}})Chen, Zhu, Shen, Li, Liu, Zhang, Krishnamoorthi, Chandra, Xiong, and Elhoseiny]{chen2023minigpt}
Jun Chen, Deyao Zhu, Xiaoqian Shen, Xiang Li, Zechun Liu, Pengchuan Zhang, Raghuraman Krishnamoorthi, Vikas Chandra, Yunyang Xiong, and Mohamed Elhoseiny.
\newblock Minigpt-v2: large language model as a unified interface for vision-language multi-task learning.
\newblock \emph{arXiv preprint arXiv:2310.09478}, 2023{\natexlab{a}}.

\bibitem[Chen et~al.(2022{\natexlab{a}})Chen, Wu, Wang, Liu, Tompkins, Chen, and Wei]{chen2022beats}
Sanyuan Chen, Yu Wu, Chengyi Wang, Shujie Liu, Daniel Tompkins, Zhuo Chen, and Furu Wei.
\newblock Beats: Audio pre-training with acoustic tokenizers.
\newblock \emph{arXiv preprint arXiv:2212.09058}, 2022{\natexlab{a}}.

\bibitem[Chen et~al.(2023{\natexlab{b}})Chen, He, Guo, Zhu, Wang, Tang, and Liu]{chen2023valor}
Sihan Chen, Xingjian He, Longteng Guo, Xinxin Zhu, Weining Wang, Jinhui Tang, and Jing Liu.
\newblock Valor: Vision-audio-language omni-perception pretraining model and dataset.
\newblock \emph{arXiv preprint arXiv:2304.08345}, 2023{\natexlab{b}}.

\bibitem[Chen et~al.(2022{\natexlab{b}})Chen, Saxena, Li, Lin, Fleet, and Hinton]{chen2022unified}
Ting Chen, Saurabh Saxena, Lala Li, Tsung-Yi Lin, David~J Fleet, and Geoffrey~E Hinton.
\newblock A unified sequence interface for vision tasks.
\newblock \emph{Advances in Neural Information Processing Systems}, 35:\penalty0 31333--31346, 2022{\natexlab{b}}.

\bibitem[Chen et~al.(2022{\natexlab{c}})Chen, Wang, Changpinyo, Piergiovanni, Padlewski, Salz, Goodman, Grycner, Mustafa, Beyer, et~al.]{chen2022pali}
Xi Chen, Xiao Wang, Soravit Changpinyo, AJ Piergiovanni, Piotr Padlewski, Daniel Salz, Sebastian Goodman, Adam Grycner, Basil Mustafa, Lucas Beyer, et~al.
\newblock Pali: A jointly-scaled multilingual language-image model.
\newblock \emph{arXiv preprint arXiv:2209.06794}, 2022{\natexlab{c}}.

\bibitem[Chen et~al.(2023{\natexlab{c}})Chen, Wang, Wang, Liu, Yin, Liu, Sheng, Ouyang, Qiao, and Shao]{chen2023octavius}
Zeren Chen, Ziqin Wang, Zhen Wang, Huayang Liu, Zhenfei Yin, Si Liu, Lu Sheng, Wanli Ouyang, Yu Qiao, and Jing Shao.
\newblock Octavius: Mitigating task interference in mllms via lora-moe.
\newblock \emph{arXiv preprint arXiv:2311.02684}, 2023{\natexlab{c}}.

\bibitem[Cheng et~al.(2024)Cheng, Leng, Zhang, Xin, Li, Chen, Zhu, Zhang, Luo, Zhao, et~al.]{cheng2024videollama}
Zesen Cheng, Sicong Leng, Hang Zhang, Yifei Xin, Xin Li, Guanzheng Chen, Yongxin Zhu, Wenqi Zhang, Ziyang Luo, Deli Zhao, et~al.
\newblock Videollama 2: Advancing spatial-temporal modeling and audio understanding in video-llms.
\newblock \emph{arXiv preprint arXiv:2406.07476}, 2024.

\bibitem[Chiang et~al.(2023)Chiang, Li, Lin, Sheng, Wu, Zhang, Zheng, Zhuang, Zhuang, Gonzalez, et~al.]{chiang2023vicuna}
Wei-Lin Chiang, Zhuohan Li, Zi Lin, Ying Sheng, Zhanghao Wu, Hao Zhang, Lianmin Zheng, Siyuan Zhuang, Yonghao Zhuang, Joseph~E Gonzalez, et~al.
\newblock Vicuna: An open-source chatbot impressing gpt-4 with 90\%* chatgpt quality.
\newblock \emph{See https://vicuna. lmsys. org (accessed 14 April 2023)}, 2\penalty0 (3):\penalty0 6, 2023.

\bibitem[Chowdhery et~al.(2023)Chowdhery, Narang, Devlin, Bosma, Mishra, Roberts, Barham, Chung, Sutton, Gehrmann, et~al.]{chowdhery2023palm}
Aakanksha Chowdhery, Sharan Narang, Jacob Devlin, Maarten Bosma, Gaurav Mishra, Adam Roberts, Paul Barham, Hyung~Won Chung, Charles Sutton, Sebastian Gehrmann, et~al.
\newblock Palm: Scaling language modeling with pathways.
\newblock \emph{Journal of Machine Learning Research}, 24\penalty0 (240):\penalty0 1--113, 2023.

\bibitem[Chowdhury et~al.(2024)Chowdhury, Nag, Dasgupta, Chen, Elhoseiny, Gao, and Manocha]{chowdhury2024meerkat}
Sanjoy Chowdhury, Sayan Nag, Subhrajyoti Dasgupta, Jun Chen, Mohamed Elhoseiny, Ruohan Gao, and Dinesh Manocha.
\newblock Meerkat: Audio-visual large language model for grounding in space and time.
\newblock \emph{arXiv preprint arXiv:2407.01851}, 2024.

\bibitem[Gao et~al.(2024)Gao, Chen, Chen, Wang, and Lu]{gao2024avsegformer}
Shengyi Gao, Zhe Chen, Guo Chen, Wenhai Wang, and Tong Lu.
\newblock Avsegformer: Audio-visual segmentation with transformer.
\newblock In \emph{Proceedings of the AAAI Conference on Artificial Intelligence}, pages 12155--12163, 2024.

\bibitem[Gupta et~al.(2019)Gupta, Dollar, and Girshick]{gupta2019lvis}
Agrim Gupta, Piotr Dollar, and Ross Girshick.
\newblock Lvis: A dataset for large vocabulary instance segmentation.
\newblock In \emph{Proceedings of the IEEE/CVF conference on computer vision and pattern recognition}, pages 5356--5364, 2019.

\bibitem[He et~al.(2024)He, Wang, Wang, Lu, He, Lan, Luo, and Xie]{he2024multi}
Junwen He, Yifan Wang, Lijun Wang, Huchuan Lu, Jun-Yan He, Jin-Peng Lan, Bin Luo, and Xuansong Xie.
\newblock Multi-modal instruction tuned llms with fine-grained visual perception.
\newblock In \emph{Proceedings of the IEEE/CVF Conference on Computer Vision and Pattern Recognition}, pages 13980--13990, 2024.

\bibitem[Hou et~al.(2024)Hou, Li, Tian, and Hu]{hou2024toward}
Wenxuan Hou, Guangyao Li, Yapeng Tian, and Di Hu.
\newblock Toward long form audio-visual video understanding.
\newblock \emph{ACM Transactions on Multimedia Computing, Communications and Applications}, 20\penalty0 (9):\penalty0 1--26, 2024.

\bibitem[Hu et~al.(2020)Hu, Qian, Jiang, Tan, Wen, Ding, Lin, and Dou]{hu2020discriminative}
Di Hu, Rui Qian, Minyue Jiang, Xiao Tan, Shilei Wen, Errui Ding, Weiyao Lin, and Dejing Dou.
\newblock Discriminative sounding objects localization via self-supervised audiovisual matching.
\newblock \emph{Advances in Neural Information Processing Systems}, 33:\penalty0 10077--10087, 2020.

\bibitem[Hu et~al.(2021)Hu, Shen, Wallis, Allen-Zhu, Li, Wang, Wang, and Chen]{hu2021lora}
Edward~J Hu, Yelong Shen, Phillip Wallis, Zeyuan Allen-Zhu, Yuanzhi Li, Shean Wang, Lu Wang, and Weizhu Chen.
\newblock Lora: Low-rank adaptation of large language models.
\newblock \emph{arXiv preprint arXiv:2106.09685}, 2021.

\bibitem[Huang et~al.(2023)Huang, Liu, Lin, Pang, Du, and Lin]{huang2023lorahub}
Chengsong Huang, Qian Liu, Bill~Yuchen Lin, Tianyu Pang, Chao Du, and Min Lin.
\newblock Lorahub: Efficient cross-task generalization via dynamic lora composition.
\newblock \emph{arXiv preprint arXiv:2307.13269}, 2023.

\bibitem[Jiang et~al.(2022)Jiang, Xu, Chen, Zhang, Song, Shen, Lu, and Shen]{jiang2022dhhn}
Xun Jiang, Xing Xu, Zhiguo Chen, Jingran Zhang, Jingkuan Song, Fumin Shen, Huimin Lu, and Heng~Tao Shen.
\newblock Dhhn: Dual hierarchical hybrid network for weakly-supervised audio-visual video parsing.
\newblock In \emph{Proceedings of the 30th ACM International Conference on Multimedia}, pages 719--727, 2022.

\bibitem[Kenton and Toutanova(2019)]{kenton2019bert}
Jacob Devlin Ming-Wei~Chang Kenton and Lee~Kristina Toutanova.
\newblock Bert: Pre-training of deep bidirectional transformers for language understanding.
\newblock In \emph{Proceedings of naacL-HLT}, page~2. Minneapolis, Minnesota, 2019.

\bibitem[Kim et~al.(2019)Kim, Kim, Lee, and Kim]{kim2019audiocaps}
Chris~Dongjoo Kim, Byeongchang Kim, Hyunmin Lee, and Gunhee Kim.
\newblock Audiocaps: Generating captions for audios in the wild.
\newblock In \emph{Proceedings of the 2019 Conference of the North American Chapter of the Association for Computational Linguistics: Human Language Technologies, Volume 1 (Long and Short Papers)}, pages 119--132, 2019.

\bibitem[Kirillov et~al.(2023)Kirillov, Mintun, Ravi, Mao, Rolland, Gustafson, Xiao, Whitehead, Berg, Lo, et~al.]{kirillov2023segment}
Alexander Kirillov, Eric Mintun, Nikhila Ravi, Hanzi Mao, Chloe Rolland, Laura Gustafson, Tete Xiao, Spencer Whitehead, Alexander~C Berg, Wan-Yen Lo, et~al.
\newblock Segment anything.
\newblock In \emph{Proceedings of the IEEE/CVF International Conference on Computer Vision}, pages 4015--4026, 2023.

\bibitem[Lai et~al.(2024)Lai, Tian, Chen, Li, Yuan, Liu, and Jia]{lai2024lisa}
Xin Lai, Zhuotao Tian, Yukang Chen, Yanwei Li, Yuhui Yuan, Shu Liu, and Jiaya Jia.
\newblock Lisa: Reasoning segmentation via large language model.
\newblock In \emph{Proceedings of the IEEE/CVF Conference on Computer Vision and Pattern Recognition}, pages 9579--9589, 2024.

\bibitem[Lao et~al.(2023)Lao, Pu, Liu, He, Bakker, and Lew]{lao2023coca}
Mingrui Lao, Nan Pu, Yu Liu, Kai He, Erwin~M Bakker, and Michael~S Lew.
\newblock Coca: Collaborative causal regularization for audio-visual question answering.
\newblock In \emph{Proceedings of the AAAI Conference on Artificial Intelligence}, pages 12995--13003, 2023.

\bibitem[Li et~al.(2022)Li, Wei, Tian, Xu, Wen, and Hu]{li2022learning}
Guangyao Li, Yake Wei, Yapeng Tian, Chenliang Xu, Ji-Rong Wen, and Di Hu.
\newblock Learning to answer questions in dynamic audio-visual scenarios.
\newblock In \emph{Proceedings of the IEEE/CVF Conference on Computer Vision and Pattern Recognition}, pages 19108--19118, 2022.

\bibitem[Li et~al.(2023{\natexlab{a}})Li, Hou, and Hu]{li2023progressive}
Guangyao Li, Wenxuan Hou, and Di Hu.
\newblock Progressive spatio-temporal perception for audio-visual question answering.
\newblock In \emph{Proceedings of the 31st ACM International Conference on Multimedia}, pages 7808--7816, 2023{\natexlab{a}}.

\bibitem[Li et~al.(2023{\natexlab{b}})Li, Xu, and Hu]{li2023multi}
Guangyao Li, Yixin Xu, and Di Hu.
\newblock Multi-scale attention for audio question answering.
\newblock \emph{arXiv preprint arXiv:2305.17993}, 2023{\natexlab{b}}.

\bibitem[Li et~al.(2024{\natexlab{a}})Li, Du, and Hu]{li2024boosting}
Guangyao Li, Henghui Du, and Di Hu.
\newblock Boosting audio visual question answering via key semantic-aware cues.
\newblock \emph{arXiv preprint arXiv:2407.20693}, 2024{\natexlab{a}}.

\bibitem[Li et~al.(2023{\natexlab{c}})Li, Li, Savarese, and Hoi]{li2023blip}
Junnan Li, Dongxu Li, Silvio Savarese, and Steven Hoi.
\newblock Blip-2: Bootstrapping language-image pre-training with frozen image encoders and large language models.
\newblock In \emph{International conference on machine learning}, pages 19730--19742. PMLR, 2023{\natexlab{c}}.

\bibitem[Li et~al.(2023{\natexlab{d}})Li, Yang, Chen, Yang, and Xiao]{li2023catr}
Kexin Li, Zongxin Yang, Lei Chen, Yi Yang, and Jun Xiao.
\newblock Catr: Combinatorial-dependence audio-queried transformer for audio-visual video segmentation.
\newblock In \emph{Proceedings of the 31st ACM International Conference on Multimedia}, pages 1485--1494, 2023{\natexlab{d}}.

\bibitem[Li et~al.(2024{\natexlab{b}})Li, Xu, Zhang, Song, Cai, Qi, Zhou, Pan, Li, Tu, et~al.]{li2024groundinggpt}
Zhaowei Li, Qi Xu, Dong Zhang, Hang Song, Yiqing Cai, Qi Qi, Ran Zhou, Junting Pan, Zefeng Li, Vu Tu, et~al.
\newblock Groundinggpt: Language enhanced multi-modal grounding model.
\newblock In \emph{Proceedings of the 62nd Annual Meeting of the Association for Computational Linguistics (Volume 1: Long Papers)}, pages 6657--6678, 2024{\natexlab{b}}.

\bibitem[Lin et~al.(2023{\natexlab{a}})Lin, Ye, Zhu, Cui, Ning, Jin, and Yuan]{lin2023video}
Bin Lin, Yang Ye, Bin Zhu, Jiaxi Cui, Munan Ning, Peng Jin, and Li Yuan.
\newblock Video-llava: Learning united visual representation by alignment before projection.
\newblock \emph{arXiv preprint arXiv:2311.10122}, 2023{\natexlab{a}}.

\bibitem[Lin and Wang(2020)]{lin2020audiovisual}
Yan-Bo Lin and Yu-Chiang~Frank Wang.
\newblock Audiovisual transformer with instance attention for audio-visual event localization.
\newblock In \emph{Proceedings of the Asian Conference on Computer Vision}, 2020.

\bibitem[Lin et~al.(2023{\natexlab{b}})Lin, Sung, Lei, Bansal, and Bertasius]{lin2023vision}
Yan-Bo Lin, Yi-Lin Sung, Jie Lei, Mohit Bansal, and Gedas Bertasius.
\newblock Vision transformers are parameter-efficient audio-visual learners.
\newblock In \emph{Proceedings of the IEEE/CVF Conference on Computer Vision and Pattern Recognition}, pages 2299--2309, 2023{\natexlab{b}}.

\bibitem[Liu et~al.(2024)Liu, Li, Zhang, Li, Huang, Wang, and Yu]{liu2024bavs}
Chen Liu, Peike Li, Hu Zhang, Lincheng Li, Zi Huang, Dadong Wang, and Xin Yu.
\newblock Bavs: bootstrapping audio-visual segmentation by integrating foundation knowledge.
\newblock \emph{IEEE Transactions on Multimedia}, 2024.

\bibitem[Lu et~al.(2022)Lu, Clark, Zellers, Mottaghi, and Kembhavi]{lu2022unified}
Jiasen Lu, Christopher Clark, Rowan Zellers, Roozbeh Mottaghi, and Aniruddha Kembhavi.
\newblock Unified-io: A unified model for vision, language, and multi-modal tasks.
\newblock In \emph{The Eleventh International Conference on Learning Representations}, 2022.

\bibitem[Lu et~al.(2024)Lu, Clark, Lee, Zhang, Khosla, Marten, Hoiem, and Kembhavi]{lu2024unified}
Jiasen Lu, Christopher Clark, Sangho Lee, Zichen Zhang, Savya Khosla, Ryan Marten, Derek Hoiem, and Aniruddha Kembhavi.
\newblock Unified-io 2: Scaling autoregressive multimodal models with vision language audio and action.
\newblock In \emph{Proceedings of the IEEE/CVF Conference on Computer Vision and Pattern Recognition}, pages 26439--26455, 2024.

\bibitem[Mao et~al.(2021)Mao, Zhang, Wan, Dai, Li, Lv, Tian, Fan, and Barnes]{mao2021transformer}
Yuxin Mao, Jing Zhang, Zhexiong Wan, Yuchao Dai, Aixuan Li, Yunqiu Lv, Xinyu Tian, Deng-Ping Fan, and Nick Barnes.
\newblock Transformer transforms salient object detection and camouflaged object detection.
\newblock \emph{arXiv preprint arXiv:2104.10127}, 1\penalty0 (2):\penalty0 5, 2021.

\bibitem[Mo and Morgado(2022)]{mo2022localizing}
Shentong Mo and Pedro Morgado.
\newblock Localizing visual sounds the easy way.
\newblock In \emph{European Conference on Computer Vision}, pages 218--234. Springer, 2022.

\bibitem[Panagopoulou et~al.(2023)Panagopoulou, Xue, Yu, Li, Li, Joty, Xu, Savarese, Xiong, and Niebles]{panagopoulou2023x}
Artemis Panagopoulou, Le Xue, Ning Yu, Junnan Li, Dongxu Li, Shafiq Joty, Ran Xu, Silvio Savarese, Caiming Xiong, and Juan~Carlos Niebles.
\newblock X-instructblip: A framework for aligning x-modal instruction-aware representations to llms and emergent cross-modal reasoning.
\newblock \emph{arXiv preprint arXiv:2311.18799}, 2023.

\bibitem[Radford et~al.(2019)Radford, Wu, Child, Luan, Amodei, Sutskever, et~al.]{radford2019language}
Alec Radford, Jeffrey Wu, Rewon Child, David Luan, Dario Amodei, Ilya Sutskever, et~al.
\newblock Language models are unsupervised multitask learners.
\newblock \emph{OpenAI blog}, 1\penalty0 (8):\penalty0 9, 2019.

\bibitem[Radford et~al.(2021)Radford, Kim, Hallacy, Ramesh, Goh, Agarwal, Sastry, Askell, Mishkin, Clark, et~al.]{radford2021learning}
Alec Radford, Jong~Wook Kim, Chris Hallacy, Aditya Ramesh, Gabriel Goh, Sandhini Agarwal, Girish Sastry, Amanda Askell, Pamela Mishkin, Jack Clark, et~al.
\newblock Learning transferable visual models from natural language supervision.
\newblock In \emph{International conference on machine learning}, pages 8748--8763. PMLR, 2021.

\bibitem[Ren et~al.(2024{\natexlab{a}})Ren, Yao, Li, Sun, and Hou]{ren2024timechat}
Shuhuai Ren, Linli Yao, Shicheng Li, Xu Sun, and Lu Hou.
\newblock Timechat: A time-sensitive multimodal large language model for long video understanding.
\newblock In \emph{Proceedings of the IEEE/CVF Conference on Computer Vision and Pattern Recognition}, pages 14313--14323, 2024{\natexlab{a}}.

\bibitem[Ren et~al.(2024{\natexlab{b}})Ren, Huang, Wei, Zhao, Fu, Feng, and Jin]{ren2024pixellm}
Zhongwei Ren, Zhicheng Huang, Yunchao Wei, Yao Zhao, Dongmei Fu, Jiashi Feng, and Xiaojie Jin.
\newblock Pixellm: Pixel reasoning with large multimodal model.
\newblock In \emph{Proceedings of the IEEE/CVF Conference on Computer Vision and Pattern Recognition}, pages 26374--26383, 2024{\natexlab{b}}.

\bibitem[Senocak et~al.(2018)Senocak, Oh, Kim, Yang, and Kweon]{senocak2018learning}
Arda Senocak, Tae-Hyun Oh, Junsik Kim, Ming-Hsuan Yang, and In~So Kweon.
\newblock Learning to localize sound source in visual scenes.
\newblock In \emph{Proceedings of the IEEE Conference on Computer Vision and Pattern Recognition}, pages 4358--4366, 2018.

\bibitem[Sun et~al.(2024)Sun, Yu, Tang, Chen, Tan, Li, Lu, Ma, Wang, and Zhang]{sun2024video}
Guangzhi Sun, Wenyi Yu, Changli Tang, Xianzhao Chen, Tian Tan, Wei Li, Lu Lu, Zejun Ma, Yuxuan Wang, and Chao Zhang.
\newblock video-salmonn: Speech-enhanced audio-visual large language models.
\newblock \emph{arXiv preprint arXiv:2406.15704}, 2024.

\bibitem[Sun et~al.(2023)Sun, Zhang, Wang, Liu, Zhong, Feng, Guo, Zhang, and Barnes]{sun2023learning}
Weixuan Sun, Jiayi Zhang, Jianyuan Wang, Zheyuan Liu, Yiran Zhong, Tianpeng Feng, Yandong Guo, Yanhao Zhang, and Nick Barnes.
\newblock Learning audio-visual source localization via false negative aware contrastive learning.
\newblock In \emph{Proceedings of the IEEE/CVF Conference on Computer Vision and Pattern Recognition}, pages 6420--6429, 2023.

\bibitem[Tian et~al.(2024)Tian, Shi, Guo, Li, and Xu]{tian2024hydralora}
Chunlin Tian, Zhan Shi, Zhijiang Guo, Li Li, and Chengzhong Xu.
\newblock Hydralora: An asymmetric lora architecture for efficient fine-tuning.
\newblock \emph{arXiv preprint arXiv:2404.19245}, 2024.

\bibitem[Tian et~al.(2018)Tian, Shi, Li, Duan, and Xu]{tian2018audio}
Yapeng Tian, Jing Shi, Bochen Li, Zhiyao Duan, and Chenliang Xu.
\newblock Audio-visual event localization in unconstrained videos.
\newblock In \emph{Proceedings of the European conference on computer vision (ECCV)}, pages 247--263, 2018.

\bibitem[Tian et~al.(2020)Tian, Li, and Xu]{tian2020unified}
Yapeng Tian, Dingzeyu Li, and Chenliang Xu.
\newblock Unified multisensory perception: Weakly-supervised audio-visual video parsing.
\newblock In \emph{Computer Vision--ECCV 2020: 16th European Conference, Glasgow, UK, August 23--28, 2020, Proceedings, Part III 16}, pages 436--454. Springer, 2020.

\bibitem[Touvron et~al.(2023)Touvron, Lavril, Izacard, Martinet, Lachaux, Lacroix, Rozi{\`e}re, Goyal, Hambro, Azhar, et~al.]{touvron2023llama}
Hugo Touvron, Thibaut Lavril, Gautier Izacard, Xavier Martinet, Marie-Anne Lachaux, Timoth{\'e}e Lacroix, Baptiste Rozi{\`e}re, Naman Goyal, Eric Hambro, Faisal Azhar, et~al.
\newblock Llama: Open and efficient foundation language models.
\newblock \emph{arXiv preprint arXiv:2302.13971}, 2023.

\bibitem[Wang et~al.(2022)Wang, Xie, Li, Fan, Song, Liang, Lu, Luo, and Shao]{wang2022pvt}
Wenhai Wang, Enze Xie, Xiang Li, Deng-Ping Fan, Kaitao Song, Ding Liang, Tong Lu, Ping Luo, and Ling Shao.
\newblock Pvt v2: Improved baselines with pyramid vision transformer.
\newblock \emph{Computational Visual Media}, 8\penalty0 (3):\penalty0 415--424, 2022.

\bibitem[Wang et~al.(2024{\natexlab{a}})Wang, Liu, Li, Ding, Hu, and Li]{wang2024prompting}
Yaoting Wang, Weisong Liu, Guangyao Li, Jian Ding, Di Hu, and Xi Li.
\newblock Prompting segmentation with sound is generalizable audio-visual source localizer.
\newblock In \emph{Proceedings of the AAAI Conference on Artificial Intelligence}, pages 5669--5677, 2024{\natexlab{a}}.

\bibitem[Wang et~al.(2024{\natexlab{b}})Wang, Sun, Zhou, Li, Zhang, and Hu]{wang2024ref}
Yaoting Wang, Peiwen Sun, Dongzhan Zhou, Guangyao Li, Honggang Zhang, and Di Hu.
\newblock Ref-avs: Refer and segment objects in audio-visual scenes.
\newblock \emph{arXiv preprint arXiv:2407.10957}, 2024{\natexlab{b}}.

\bibitem[Xia and Zhao(2022)]{xia2022cross}
Yan Xia and Zhou Zhao.
\newblock Cross-modal background suppression for audio-visual event localization.
\newblock In \emph{Proceedings of the IEEE/CVF conference on computer vision and pattern recognition}, pages 19989--19998, 2022.

\bibitem[Yang et~al.(2022)Yang, Gan, Wang, Hu, Ahmed, Liu, Lu, and Wang]{yang2022unitab}
Zhengyuan Yang, Zhe Gan, Jianfeng Wang, Xiaowei Hu, Faisal Ahmed, Zicheng Liu, Yumao Lu, and Lijuan Wang.
\newblock Unitab: Unifying text and box outputs for grounded vision-language modeling.
\newblock In \emph{European Conference on Computer Vision}, pages 521--539. Springer, 2022.

\bibitem[Yu et~al.(2021)Yu, Cheng, and Feng]{yu2021mpn}
Jiashuo Yu, Ying Cheng, and Rui Feng.
\newblock Mpn: Multimodal parallel network for audio-visual event localization.
\newblock In \emph{2021 IEEE International Conference on Multimedia and Expo (ICME)}, pages 1--6. IEEE, 2021.

\bibitem[Yu et~al.(2022)Yu, Cheng, Zhao, Feng, and Zhang]{yu2022mm}
Jiashuo Yu, Ying Cheng, Rui-Wei Zhao, Rui Feng, and Yuejie Zhang.
\newblock Mm-pyramid: Multimodal pyramid attentional network for audio-visual event localization and video parsing.
\newblock In \emph{Proceedings of the 30th ACM international conference on multimedia}, pages 6241--6249, 2022.

\bibitem[Zeng et~al.(2022)Zeng, Liu, Du, Wang, Lai, Ding, Yang, Xu, Zheng, Xia, et~al.]{zeng2022glm}
Aohan Zeng, Xiao Liu, Zhengxiao Du, Zihan Wang, Hanyu Lai, Ming Ding, Zhuoyi Yang, Yifan Xu, Wendi Zheng, Xiao Xia, et~al.
\newblock Glm-130b: An open bilingual pre-trained model.
\newblock \emph{arXiv preprint arXiv:2210.02414}, 2022.

\bibitem[Zhang et~al.(2023)Zhang, Li, and Bing]{zhang2023video}
Hang Zhang, Xin Li, and Lidong Bing.
\newblock Video-llama: An instruction-tuned audio-visual language model for video understanding.
\newblock \emph{arXiv preprint arXiv:2306.02858}, 2023.

\bibitem[Zhang et~al.(2021)Zhang, Xie, Barnes, and Li]{zhang2021learning}
Jing Zhang, Jianwen Xie, Nick Barnes, and Ping Li.
\newblock Learning generative vision transformer with energy-based latent space for saliency prediction.
\newblock \emph{Advances in Neural Information Processing Systems}, 34:\penalty0 15448--15463, 2021.

\bibitem[Zhou et~al.(2021)Zhou, Zheng, Zhong, Hao, and Wang]{zhou2021positive}
Jinxing Zhou, Liang Zheng, Yiran Zhong, Shijie Hao, and Meng Wang.
\newblock Positive sample propagation along the audio-visual event line.
\newblock In \emph{Proceedings of the IEEE/CVF Conference on Computer Vision and Pattern Recognition}, pages 8436--8444, 2021.

\bibitem[Zhou et~al.(2022)Zhou, Wang, Zhang, Sun, Zhang, Birchfield, Guo, Kong, Wang, and Zhong]{zhou2022audio}
Jinxing Zhou, Jianyuan Wang, Jiayi Zhang, Weixuan Sun, Jing Zhang, Stan Birchfield, Dan Guo, Lingpeng Kong, Meng Wang, and Yiran Zhong.
\newblock Audio--visual segmentation.
\newblock In \emph{European Conference on Computer Vision}, pages 386--403. Springer, 2022.

\end{thebibliography}
}

\appendix
\clearpage
\setcounter{page}{1}
\maketitlesupplementary







%


\def\paperID{6963} 
\def\confName{CVPR}
\def\confYear{2025}

\section{Dataset Construction}
In this section, we introduce the prompt used in the construction of our AV-UIE dataset and some examples in dataset. We have provided the code and annotation information for the dataset in the attachment. The audio and video data can be downloaded from these links: \href{https://github.com/YapengTian/AVE-ECCV18}{AVE}, \href{https://github.com/YapengTian/AVVP-ECCV20}{AVVP}, \href{https://github.com/OpenNLPLab/AVSBench}{ARIG}, \href{https://github.com/OpenNLPLab/AVSBench}{AVS}, \href{https://github.com/GeWu-Lab/Ref-AVS}{Ref-AVS}, \href{https://github.com/GeWu-Lab/MUSIC-AVQA}{MUSIC-AVQA}, \href{https://github.com/TXH-mercury/VALOR}{VALOR}.

\subsection{Prompt Template}
\begin{figure*}[t]
     \centering
     \includegraphics[width=0.925\textwidth]{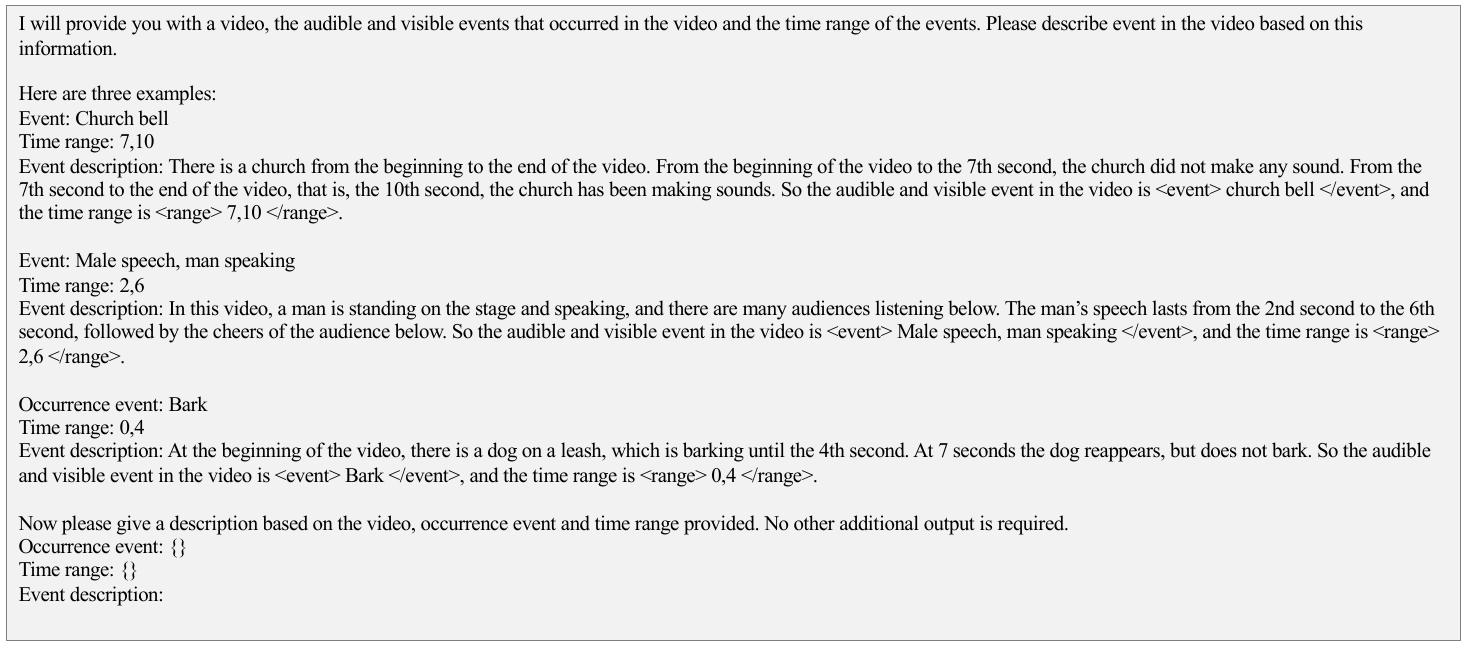}
     \vspace{-0.5em}
     \caption{
     The prompt used to convert labels on AVE task.
     }
     \label{fig:ave-prompt}
\end{figure*}

\begin{figure*}[t]
     \centering
     \includegraphics[width=0.925\textwidth]{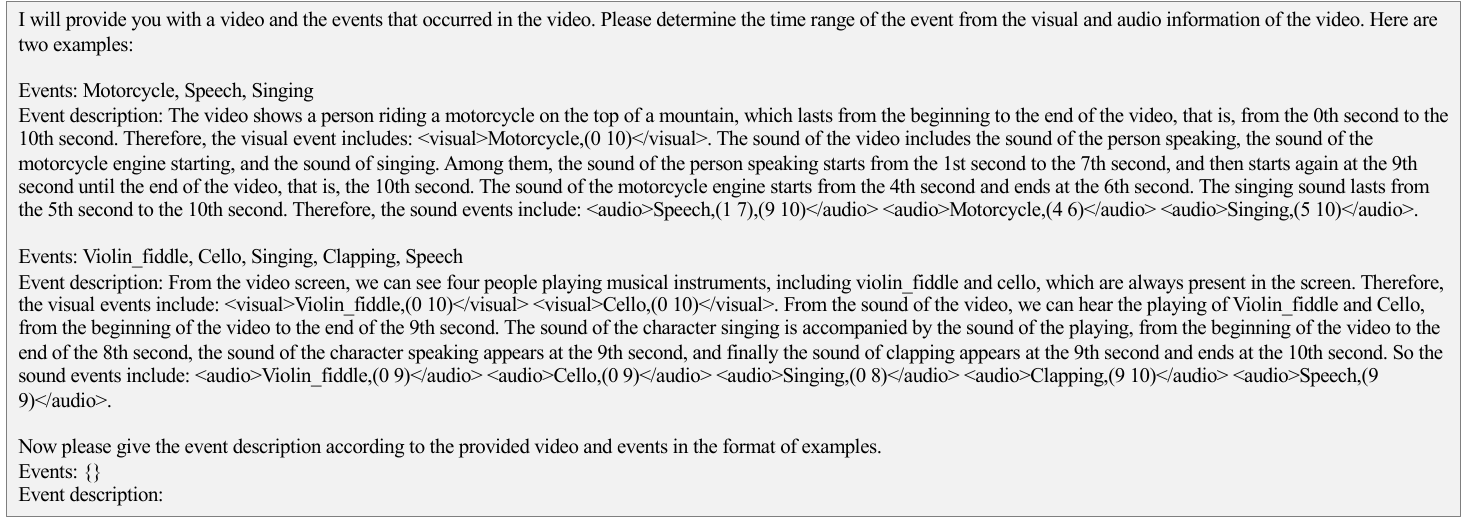}
     \vspace{-0.5em}
     \caption{
     The prompt used to convert labels on AVVP task.
     }
     \label{fig:avvp-prompt}
     \vspace{-1em}
\end{figure*}

\begin{figure*}[t]
     \centering
     \includegraphics[width=0.925\textwidth]{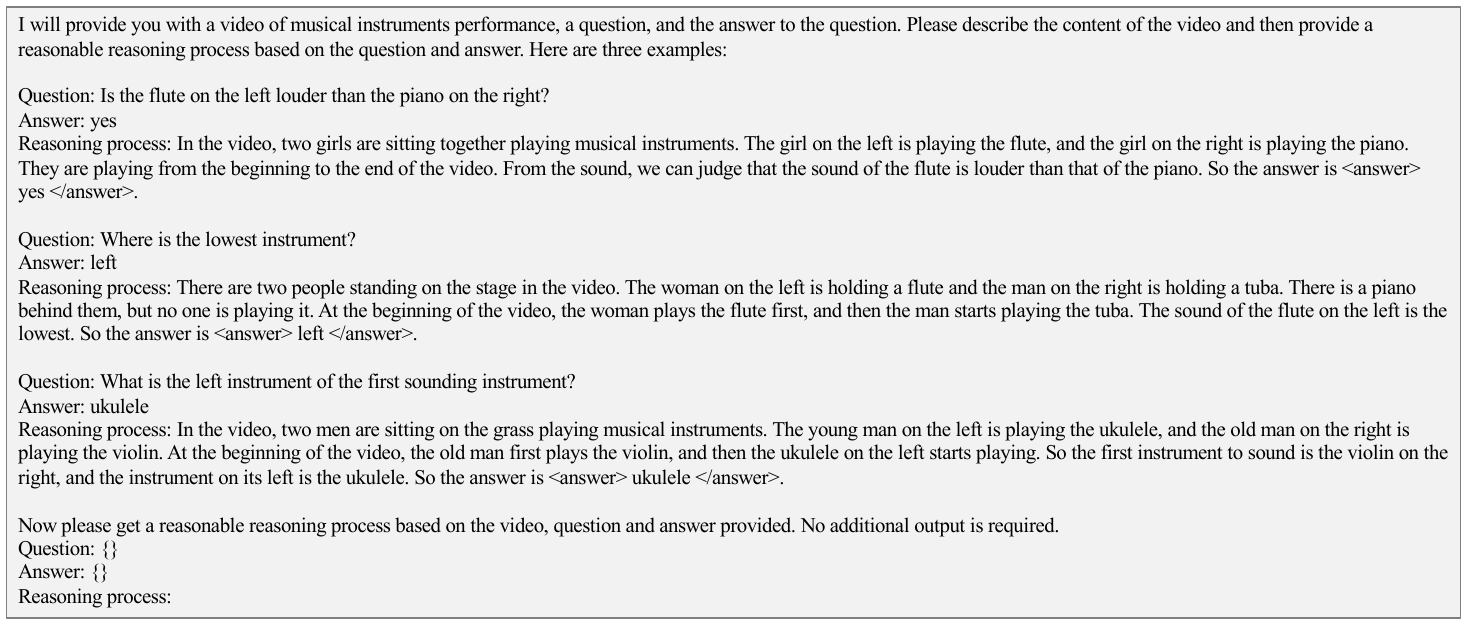}
     \vspace{-0.5em}
     \caption{
     The prompt used to convert labels on AVQA task.
     }
     \label{fig:avqa-prompt}
     \vspace{-1em}
\end{figure*}

To construct the AV-UIE dataset, we use the in-context learning approach to prompt Gemini 1.5 Pro to transform simple labels into explicit reasoning process. Fig.~\ref{fig:ave-prompt}, Fig.~\ref{fig:avvp-prompt} and Fig.~\ref{fig:avqa-prompt} demonstrate the prompt for AVE, AVVP, and AVQA tasks respectively. Specifically, for each instance in the dataset, our prompt first includes several input-output pairs, allowing the Gemini 1.5 Pro to generate content in a fixed format based on these examples. Subsequently, we provide audio, video, and simple labels to Gemini 1.5 Pro, which then outputs the reasoning process based on the these provided information, thereby clarifying cooperation relationship among audio-visual tasks. Throughout this process, we ensure that the transformed labels remain consistent with the original ones.



\subsection{Dataset Examples}
To help readers better understand the format of our dataset, we introduce a few examples below.

\begin{figure*}[t]
  \centering
     \includegraphics[width=0.99\textwidth]{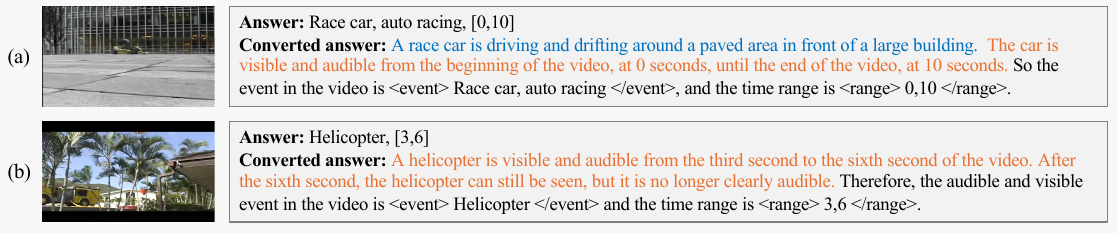}
     \vspace{-0.5em}
     \caption{
     Two examples on AVE task. (a) A race car is driving in front of a building. (b) A scene of a helicopter flying.
     }
     \label{fig:ave-example}
     \vspace{-0.5em}
\end{figure*}

\textbf{AVE task.} Fig.~\ref{fig:ave-example} are two examples on AVE task in our dataset. The video in Fig.~\ref{fig:ave-example}(a) shows a scene of a race car driving in an open space in front of a building, which lasts from the beginning to the end of the video, so the original label is \textit{``Race car, auto racing, [0,10]"}. As can be seen, in our dataset, the transformed labels include an explicit reasoning process, which is conducive to clarify cooperative relationship among tasks.

\begin{figure*}[t]
  \centering
     \includegraphics[width=0.99\textwidth]{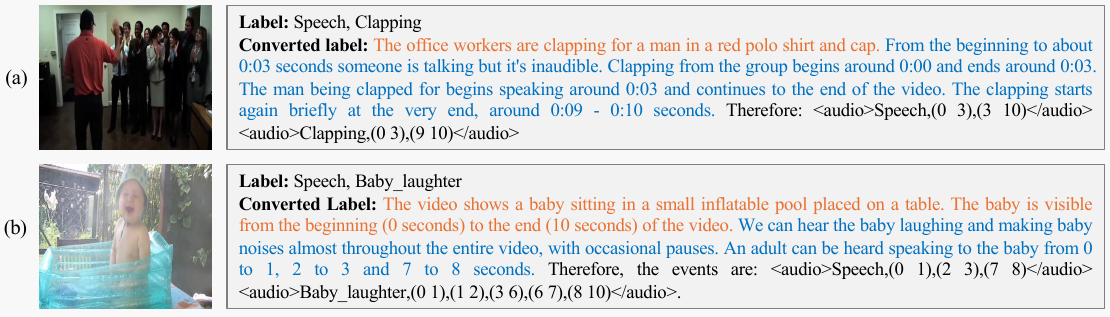}
     \vspace{-0.5em}
     \caption{
     Two examples of label transformation on AVVP task. (a) A scene of office workers clapping and someone talking. (b) A scene of a child playing and laughing in the water.
     }
     \label{fig:avvp-example}
     \vspace{-0.5em}
\end{figure*}

\textbf{AVVP task.} Fig.~\ref{fig:avvp-example} shows the label transformation results of the LLP dataset. Since there are only video-level event labels in the training set of LLP, such as \textit{``Speech, Baby\_laughter"} shown in the Fig.~\ref{fig:avvp-example}(b), but no specific time boundaries for the occurrence of events. We further annotate the temporal information of these events using Gemini 1.5 Pro and output the explicit reasoning process.

\begin{figure}[t]
  \centering
     \includegraphics[width=0.48\textwidth]{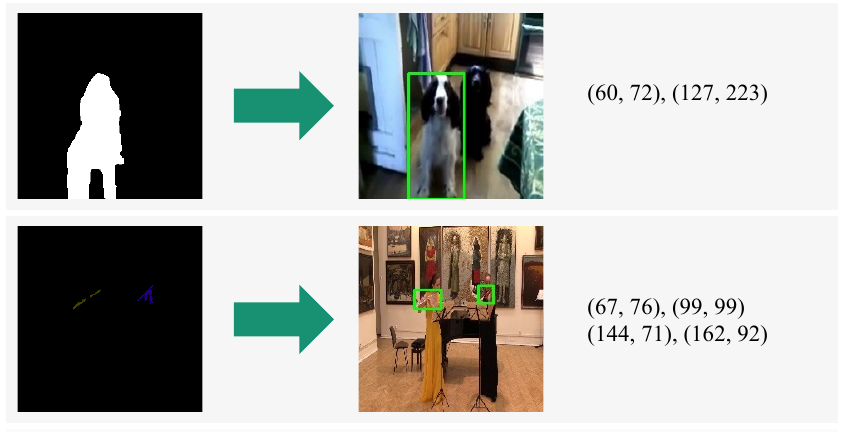}
     \vspace{-1em}
     \caption{
     Two examples of converting segmentation mask to obtain bounding boxes on AVS-Bench dataset. (a) The example on S4 subset. (b) The example on AVSS subset.
     }
     \label{fig:arig-example}
     \vspace{-0.5em}
\end{figure}

\textbf{ARIG task.} As shown in Fig.~\ref{fig:arig-example}, we convert the segmentation mask in the AVS-Bench training set into bounding boxes, then obtain the two vertices at the top left and bottom right, and represent them in the form of [$x_{Left}, y_{Top}, x_{Right}, y_{Bottom}$].

\begin{figure}[t]
  \centering
  \vspace{-0.1em}
     \includegraphics[width=0.48\textwidth]{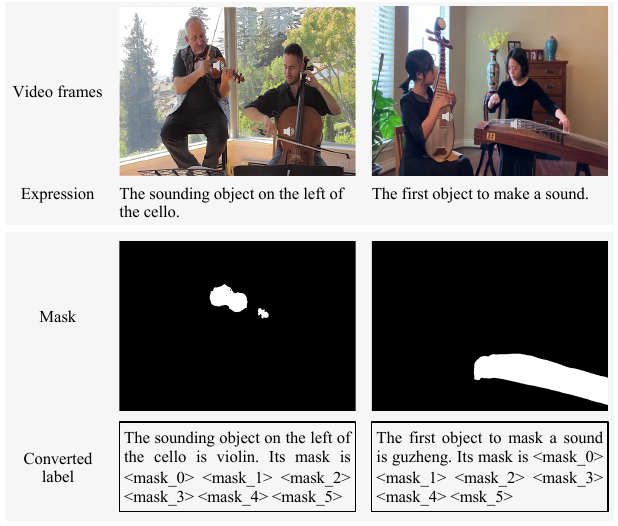}
     \vspace{-1em}
     \caption{
     Two examples of label transformation on Ref-AVS task.
     }
     \label{fig:ref-avs-example}
     \vspace{-1em}
\end{figure}

\textbf{AVS task.} As shown in Fig.~\ref{fig:ref-avs-example}, the transformed labels additionally contain temporal and spatial information. The model first needs to accurately understand the audiovisual scene, then point out the target object, and finally predict the segmentation mask based on these information.

\begin{figure*}[t]
  \centering
     \includegraphics[width=0.99\textwidth]{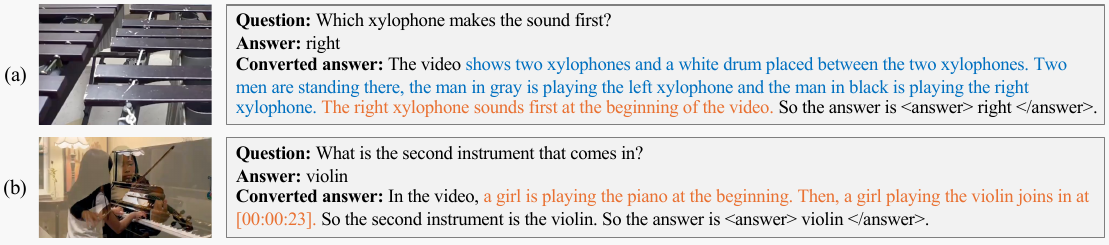}
     \vspace{-1em}
     \caption{
     Two examples of label transformation on AVQA task.
     }
     \label{fig:avqa-example}
\end{figure*}

\textbf{AVQA task.} Fig.~\ref{fig:avqa-example} shows the results after label transformation on MUSIC-AVQA dataset. It can be found that the transformed explicit reasoning process contains spatio-temporal reasoning information, such as the localization of instruments, the accurate time of sound, and the order of instruments occurrence, \textit{etc}. These information can help the model establish relationships and achieve mutual cooperation among tasks.

\subsection{Explicit Cooperation among Tasks}
\begin{figure*}[t]
    \centering
    \vspace{-2em}
    \includegraphics[width=0.85\textwidth]{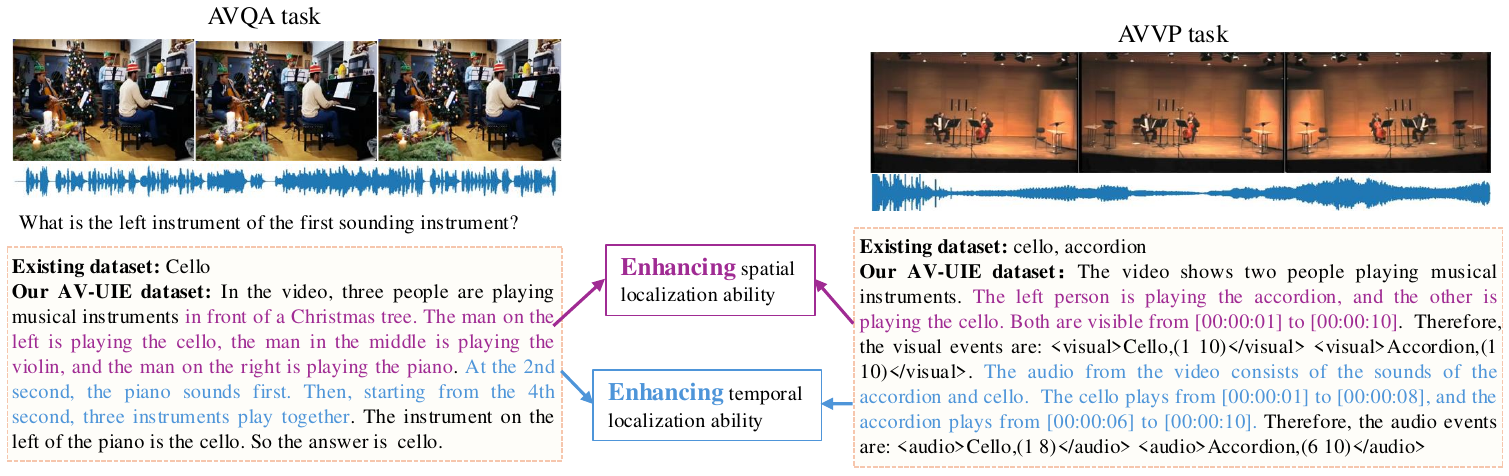}
    \vspace{-0.5em}
    \caption{AVQA and AVVP tasks achieve explicit cooperation through explicit reasoning process.}
    \label{fig:dataset-supp}
    \vspace{-1.5em}
\end{figure*}

To further illustrate how our AV-UIE dataset achieves explicit cooperation among tasks, we provide two data examples in Fig~\ref{fig:dataset-supp}. Existing audiovisual datasets only have single words ``cello, accordion". In our AV-UIE dataset, explicit reasoning process includes \textit{temporal and spatial} information (marked with different colors in Fig~\ref{fig:dataset-supp}), which can effectively achieve explicit cooperation among tasks. For example, in AVVP task, there are specific timestamps, which can enhance model's temporal localization ability, and thus help AVQA task. This is also a major difference between our dataset and existing instruction-tuning datasets, which do not emphasize concrete temporal and spatial information to achieve task cooperation.

\section{Mask decoder}

\begin{figure*}[t]
  \centering
     \includegraphics[width=0.99\textwidth]{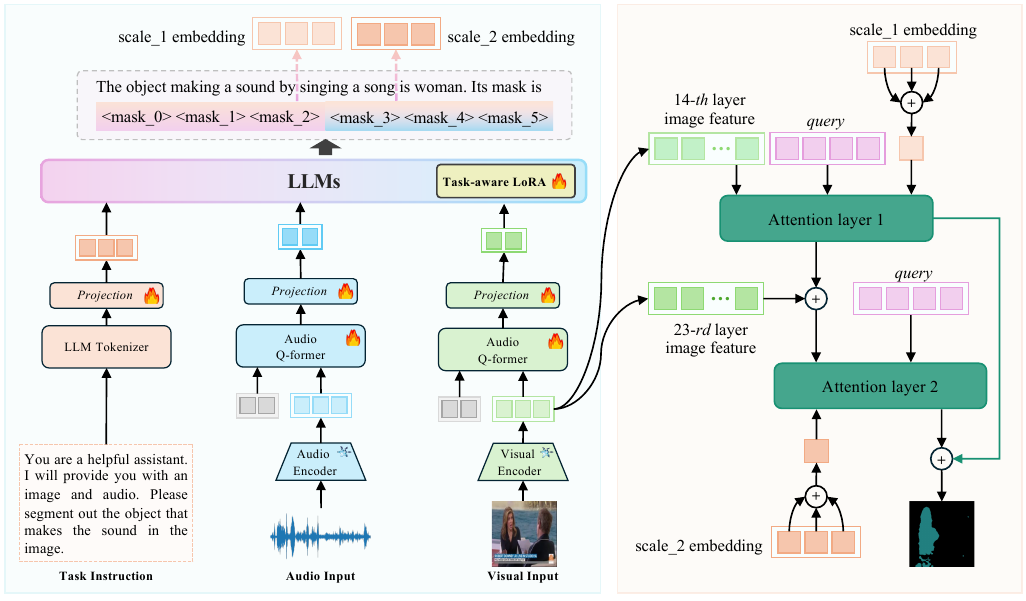}
     \vspace{-0.5em}
     \caption{
     The overview of mask decoder.
     }
     \label{fig:mask-decoder}
     \vspace{-1em}
\end{figure*}

Fig.~\ref{fig:mask-decoder} illustrates the detailed process of the mask decoder outputting the segmentation mask. As mentioned in the main paper, we obtain the LLM last-layer hidden states corresponding to the \texttt{<mask>} tokens. Inspired by the multi-scale features of PVT-v2~\cite{wang2022pvt}, these hidden states are divided into two scales. Then, we use learnable weighting factors to combine all the hidden states from each scale, resulting in two embeddings. Each embedding serves as a prompt and is sent to the mask decoder along with the corresponding scale of visual features. The mask decoder primarily consists of two cross-attention blocks, each block responding for the interaction at one scale. For each mask generation, mask decoder sequentially produces a mask score map, which then guides the model to focus attention on areas of higher relevance at the next scale, thereby promoting more accurate mask generation. Finally, we use learnable weighting factors to combine the mask maps from all scales to obtain the final segmentation mask.

\section{Experiment Results}
\begin{table*}[ht]
\begin{center}
\caption{The comparison results with specific models on AVS-Bench and Ref-AVS test set. S4, MS3 and AVSS are the subtasks of AVS-Bench. Seen, Unseen and Null are the subtasks of Ref-AVS test set.}
\vspace{-0.5em}
\label{supp_tab_avs_ref_avs}
\scalebox{0.88}{
\begin{tabular}{c|c|cc|cc|cc|cc|cc|c}
\hline
\multirow{2}{*}{\textbf{Method}} & 
\multicolumn{1}{c|}{\textbf{Visual}} & 
\multicolumn{2}{c|}{\textbf{S4}} &         
\multicolumn{2}{c|}{\textbf{MS3}} &         
\multicolumn{2}{c|}{\textbf{AVSS}} &
\multicolumn{2}{c|}{\textbf{Seen}} &
\multicolumn{2}{c|}{\textbf{Unseen}} & 
\textbf{Null}
\\
& \textbf{Backbone} & \textbf{miou} & \textbf{F-score} & \textbf{miou}  & \textbf{F-score} & \textbf{miou} & \textbf{F-score} & \textbf{miou} & \textbf{F-score} & \textbf{miou} & \textbf{F-score} & \textbf{S($\downarrow$)} \\
\hline
TPAVI~\cite{zhou2022audio} & ResNet-50 & 72.80 & 84.80 & 47.90 & 57.80 & 20.20 & 25.20 & - & - & - & - & - \\
CATR~\cite{li2023catr} & ResNet-50 & 74.80 & 86.60 & 52.80 & 65.30 & - & -  & - & - & - & - & -\\
BAVS~\cite{liu2024bavs} & ResNet-50 & 78.00 & 85.30 & 50.20 & 62.40  & 24.70 & 29.60 & - & - & - & - & - \\
\hline
iGAN~\cite{mao2021transformer} & Swin-T & 61.60 & 77.80 & 42.90 & 54.40 & - & - & - & - & - & - & - \\
LGVT~\cite{zhang2021learning} & Swin-T & 74.90 & 87.30 & 40.70 & 59.30 & - & - & - & - & - & - & - \\
\hline
TPAVI~\cite{zhou2022audio} & PVT-v2 & \underline{78.70} & \underline{87.90} & 54.00 & 64.50 & \textbf{29.80} & \textbf{35.20}  & - & - & - & - & - \\
AVSBench~\cite{zhou2022audio} & PVT-v2 & \underline{78.70} & \underline{87.90} & 54.00 & 23.20 & - & - & 0.51 & 32.36 & 0.55  & 0.21 & 0.21 \\
AVSegFormer~\cite{gao2024avsegformer} & PVT-v2 & \textbf{82.10} & \textbf{89.90} & \textbf{58.40} & \textbf{69.30} & 24.90 & 29.30 & 33.47 & 0.47 & 36.05 & 0.50  & 0.17 \\
GAVS~\cite{wang2024prompting} & PVT-v2 & - & - & - & - & - & - & 28.93 & 0.50 & 29.82 & 0.50 & 0.19  \\
EEMC~\cite{wang2024ref} & PVT-v2 & - & - & - & - & - & - & \underline{34.20} & \underline{0.51} & \textbf{49.54} & \textbf{0.65} & \textbf{0.01} \\ 
\hline
\textbf{Crab(Ours)} & \textbf{ViT/L-14} & 73.25 & 86.81 & \underline{58.21} & \underline{66.24} & \underline{26.59} & \underline{32.10} & \textbf{40.54} & \textbf{0.58} & \underline{45.55} & \underline{0.63} & \textbf{0.01} \\ 
\hline
\end{tabular}
}
\end{center}
\vspace{-1em}
\end{table*}

\begin{table*}[ht]
\begin{center}
\caption{The comparison results with specific models on MUSIC-AVQA test set.}
\vspace{-0.5em}
\label{supp_tab_avqa}
\scalebox{0.91}{
\begin{tabular}{c|ccc|ccc|cccccc|c}
\hline
& \multicolumn{3}{c|}{\textbf{Audio}}  & \multicolumn{3}{c|}{\textbf{Visual}}   & \multicolumn{6}{c|}{\textbf{Audio-Visual}}  &  \\
\multirow{-2}{*}{\textbf{Method}}  & \textbf{Count} & \textbf{Comp}  & \textbf{Avg}   & \textbf{Count} & \textbf{Local} & \textbf{Avg}   & \textbf{Exist} & \textbf{Count} & \textbf{Local} & \textbf{Comp}  & \textbf{Temp}  & \textbf{Avg}   & \multirow{-2}{*}{\textbf{Avg}} \\ \hline


ST-AVQA~\cite{li2022learning} & 77.78 & \underline{67.17} & 73.87 & 73.52 & 75.27 & 74.40 & 82.49 & 69.88 & 64.24 & \textbf{64.67} & 65.82 & 69.53 & 71.59  \\
COCA~\cite{lao2023coca} & 79.35 & \textbf{67.68} & 75.42 & 75.10 & 75.43 & 75.23 & \textbf{83.50} & 66.63 & 69.72 & 64.12 & 65.57 & 69.96 & 72.33  \\
PSTP-Net~\cite{li2023progressive} & 73.97 & 65.59 & 70.91 & 77.15 & 77.36 & 77.26 & 76.18 & 72.23 & 71.80 & 71.79 & 69.00 & 72.57 & 73.52 \\
LAVISH~\cite{lin2023vision} & 82.09 & 65.56 & 75.97 & 78.98 & 81.43 & 80.22 & 81.71 & 75.51 & 66.13 & 63.77 & 67.96 & 71.26 & 74.46 \\
TSPM~\cite{li2024boosting} & \underline{84.07} & 64.65 & \textbf{76.91} & \underline{82.29} & \underline{84.90} & \underline{83.61} & 82.19 & \underline{76.21} & \underline{71.85} & \textbf{65.76} & \textbf{71.17} & \underline{73.51} & \underline{76.79} \\
\hline

\textbf{Crab(Ours)} & \textbf{85.55} & 61.21 & \underline{76.58} & \textbf{87.51} & \textbf{93.92} & \textbf{90.73} & \underline{82.88} & \textbf{81.26} & \textbf{71.95} & 62.13 & \underline{71.11} & \textbf{74.13} & \textbf{78.94} \\ 
\hline
\end{tabular}
}
\end{center}
\vspace{-1em}
\end{table*}

\begin{table*}[t]
\begin{center}
\caption{The ablation results on task-aware LoRA. ``three heads" means the head numbers of task-aware LoRA is three.}
\vspace{-0.5em}
\label{supp_tab_ablation_lora}
\scalebox{0.775}{
\begin{tabular}{c|c|cc|c|cc|cc|cc|cc|cc|c}
\hline
\multirow{2}{*}{\textbf{Method}} & \multicolumn{1}{c|}{\textbf{AVE}} & \multicolumn{2}{c|}{\textbf{ARIG}} & \multicolumn{1}{c|}{\textbf{AVQA}} & \multicolumn{2}{c|}{\textbf{S4}} & \multicolumn{2}{c|}{\textbf{MS3}} & \multicolumn{2}{c|}{\textbf{AVSS}} & \multicolumn{2}{c}{\textbf{Seen}} & \multicolumn{2}{c|}{\textbf{Unseen}} & \multicolumn{1}{c}{\textbf{Null}} \\
& \textbf{Acc} & \textbf{cIoU} & \textbf{AUC} & \textbf{Acc} & \textbf{mIoU} & \textbf{F-score} & \textbf{mIoU} & \textbf{F-score} & \textbf{mIoU} & \textbf{F-score} & \textbf{mIoU} & \textbf{F-score} & \textbf{mIoU} & \textbf{F-score} & \textbf{S$\downarrow$} \\
\hline
\textbf{three heads} & 80.15 & 41.78 & 0.42 & \textbf{78.94} & 73.52 & 86.81 & 58.21 & 66.24 & \textbf{26.59} & \textbf{32.10} & 40.54 & \textbf{0.58} & \textbf{45.55} & \textbf{0.63} & \textbf{0.01} \\
\textbf{four heads} & \textbf{80.25} & 42.33 & 0.42 & 77.76 &  72.59 & 86.21 & 56.29 & 65.63 & 24.76 & 30.25 & \textbf{42.17} & \textbf{0.58} & 42.01 & 0.58 &  \textbf{0.01}  \\
\textbf{five heads} & 80.17 & \textbf{43.59} & \textbf{0.44} & 78.13 & \textbf{73.78} & \textbf{86.88} & \textbf{59.97} & \textbf{68.43} & 24.58 & 30.16 & 41.98 & \textbf{0.58} & 44.13 & 0.61 & \textbf{0.01} \\
\hline
\end{tabular}
}
\end{center}
\vspace{-1.25em}
\end{table*}

\subsection{More Comprehensive Ablation Results}
Multitask interference was mentioned in previous works~\cite{chen2023octavius}. It has often been overlooked in MLLMs. We provide experiments in Tab~\ref{tab_ablation}. Comparing single task and LoRA baseline (training jointly on multitask), increasing task numbers indeed improves model's performance (AVE and AVVP), but it could also introduce the issue of interference (AVQA and ARIG).

Our interaction-aware LoRA structure is a special MoE structure, where each decoder head can be seen as an expert with specific ability. We also compared results of LoRA baseline and LoRA MoE in Tab~\ref{tab_ablation}, which can also prove effectiveness of this structure.

\subsection{Pixel-level understanding}

Tab.~\ref{supp_tab_avs_ref_avs} shows the experimental results compared with specialized models on the AVS-Bench~\cite{zhou2022audio} and Ref-AVS\cite{wang2024ref} test sets. It can be seen that our model achieves comparable results on the MS3, AVSS, and Unseen subtasks, and performed best on the Seen and Null subtasks, but performed poorly on the S4 subtask. The AVS task uses audio as guiding information to find out the target object to be segmented, while text reference expressions are used in the Ref-AVS task. Since LLMs naturally have stronger understanding and reasoning capabilities for text, the performance on the three subtasks of Ref-AVS is generally better. In addition, while the model can accurately determine the position of target object, the mask decoder is responsible for outputting the segmentation mask, which will also affect the final result on these two tasks.

\subsection{Spatio-temporal reasoning}
Tab.~\ref{supp_tab_avqa} shows detailed experiment results on MUSIC-AVQA~\cite{li2022learning} test set. It can be seen that our method outperforms all specialized models. Specifically, compared to recent TSPM~\cite{li2024boosting}, our model achieves significant overall performance improvements of 2.15\% (78.94\% \textit{vs.} 76.79\%). In all visual subtasks, including \textit{Count} and \textit{Localization}, our method acheives remarkable improvements of 5.22\% (87.51\% \textit{vs.} 82.29\%) and 9.02\% (93.92\% \textit{vs.} 84.90\%) respectively. In complex audio-visual question types, our model obtains the best overall performance (74.13\%) and the performance in audio subtasks is also comparable. It is worth noting that our model generally achieves superior results in spatial localization, counting and temporal question types. This is mainly due to the cooperation of temporal and spatial localization tasks, which also proves that our method can achieve mutual cooperation among tasks. Moreover, our model performs pooly in the \textit{Comp} question type, which mainly involves subjective questions such as comparing the melody, pitch and intensity of sounds. Differences among annotators can effect the accuracy of original labels. This also makes it difficult for our model to 
learn a unified standard for evaluation, leading to a disordered reasoning process and resulting in decreased performance.

\subsection{Ablation results on LoRA head numbers}
\label{ablation-lora}
In order to explore the impact of different numbers of LoRA heads on model performance, we compare the experimental results of using three, four, and five LoRA heads respectively. Tab.~\ref{supp_tab_ablation_lora} presents the corresponding experimental results. From the table, it can be seen that different tasks achieve the best results on different numbers of LoRA heads. The increase in the number of LoRA heads does not necessarily improve the model's performance. A possible reason is that when the number of LoRA heads is too large, exceeding the types of audiovisual data interactions, the each additional LoRA head may focus on the same or multiple aspects of data interaction. The former improves the performance of the corresponding tasks, while the latter causes the same ability to be distributed among multiple LoRA heads, making it difficult for model to acquire all the abilities to solve the corresponding tasks.

\begin{figure}[t]
  \centering
     \includegraphics[width=0.48\textwidth]{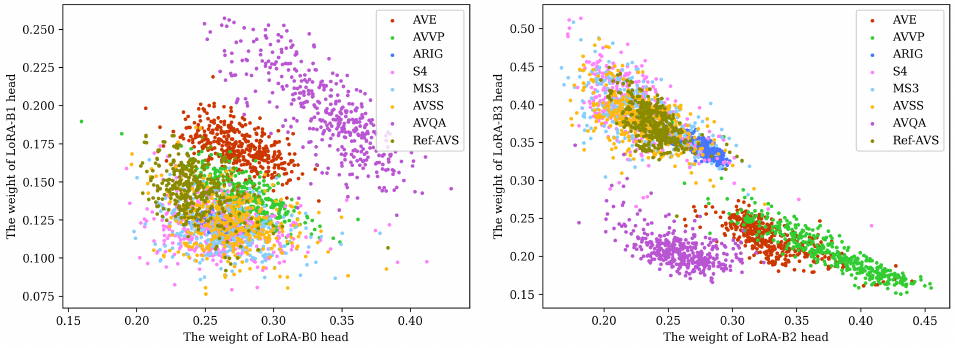}
     \vspace{-1em}
     \caption{
     The router weights of four heads.
     }
     \label{fig:router-weight-4-heads}
\end{figure}

\subsection{The visualized results on LoRA head numbers}
\begin{figure*}[t]
  \centering
     \includegraphics[width=1\textwidth]{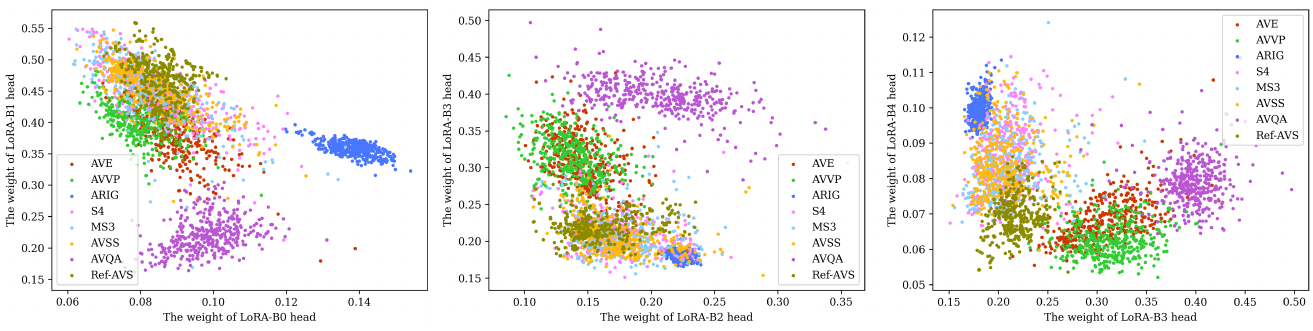}
     \vspace{-1em}
     \caption{
     The router weights of five heads.
     }
     \label{fig:router-weight-5-heads}
     \vspace{-1em}
\end{figure*}

Furthermore, we also perform a visualized analysis of the router weights for different numbers of LoRA heads. Fig.~\ref{fig:router-weight-4-heads} and Fig.~\ref{fig:router-weight-5-heads} demonstrate the router weights when using four heads and five heads. Similar to the analysis in the main paper, we can see that tasks of the same type form a cluster, indicating that their dependence on the same head is similar. Different types of tasks have different dependencies on these heads, indicating that different heads have different types of capabilities. Moreover, as discussed in Section~\ref{ablation-lora}, when the number of LoRA heads is too large, each head may focus on a specific aspect of data interaction, thus possessing a specific type of capability, such as the \textit{head-B2} and \textit{head-B3} in four heads, 
\textit{head-B3} in five heads. It may also have multiple capabilities at the same time, such as \textit{head-B0} in four heads, \textit{head-B1}, \textit{head-B2} and \textit{head-B3} in five heads. Therefore, how to more precisely control each head to have specific capability may be a meaningful direction for future research.

\subsection{Visualized results on all tasks}
\begin{figure*}[t]
  \centering
     \includegraphics[width=0.99\textwidth]{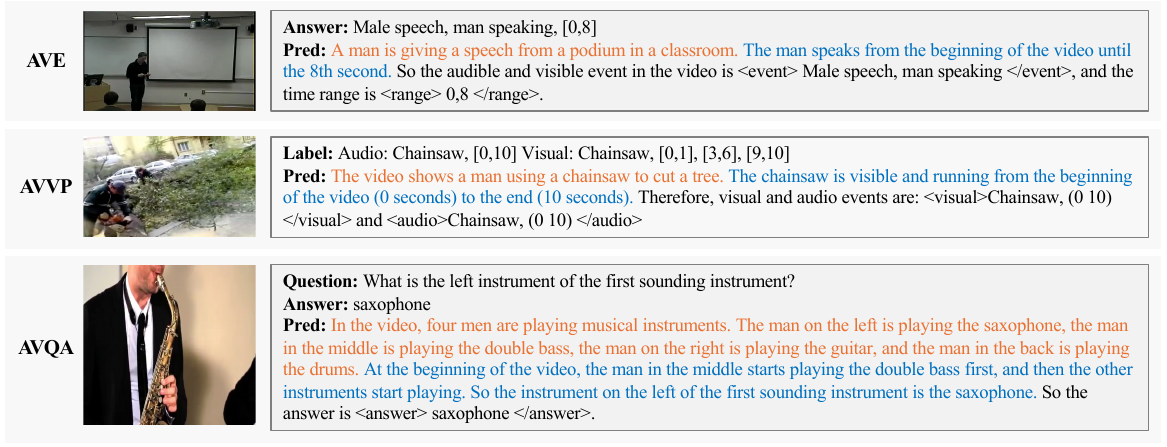}
     \vspace{-0.5em}
     \caption{
     The visualized results on AVE, AVVP and AVQA tasks.
     }
     \label{fig:vis-ave-avvp-avqa-1}
     \vspace{-1em}
\end{figure*}

\begin{figure*}[t]
  \centering
     \includegraphics[width=0.99\textwidth]{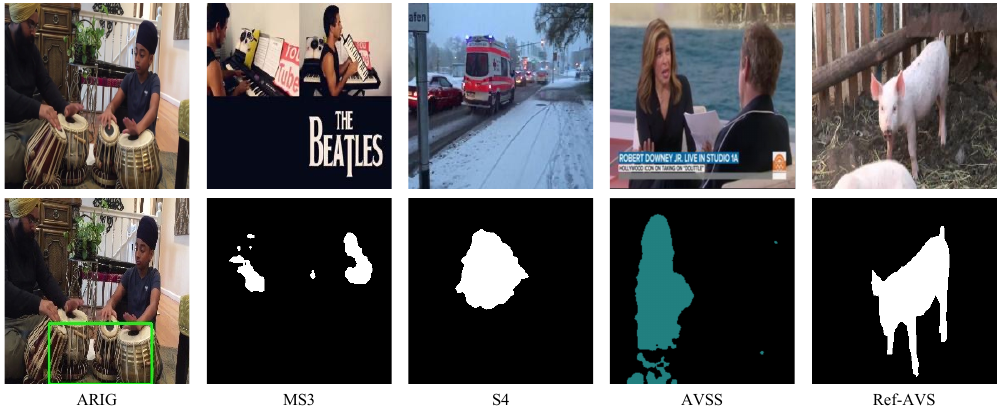}
     \vspace{-0.5em}
     \caption{
     The visualized results on ARIG, AVS and Ref-AVS tasks.
     }
     \label{fig:vis-arig-avs-1}
     \vspace{-1em}
\end{figure*}

Fig.~\ref{fig:vis-ave-avvp-avqa-1} and Fig.~\ref{fig:vis-arig-avs-1} present some visualized results on all tasks.

\begin{table*}[t]
\belowrulesep=0pt
\aboverulesep=0pt
\begin{center}
\caption{More comprehensive ablation results. ERP represents reasoning process. IA-LoRA represents interaction-aware LoRA.}
\label{tab_ablation-ab}
\vspace{-1em}
\scalebox{0.85}{
\begin{tabular}{c|c|c|cc|cc}
\toprule
\multirow{2}{*}{\textbf{Method}} & \multicolumn{1}{c|}{\textbf{AVQA}} & \multicolumn{1}{c|}{\textbf{AVE}} & \multicolumn{2}{c|}{\textbf{AVVP}} & \multicolumn{2}{c}{\textbf{ARIG}} \\
&\textbf{Avg} & \textbf{Acc} & \textbf{Segment} & \textbf{Event} & \textbf{cIoU} & \textbf{AUC} \\
\hline
Single task & 75.87 & 79.10 & 56.11 & 51.32 & 39.93 & 0.40 \\
LoRA baseline & 75.78 & 79.55 & 56.91 & 52.13 & 39.87 & 0.40 \\
LoRA MoE & 77.60 & 80.02 &  58.21 & 53.32  & 41.36 & 0.42 \\
\textit{w/o.} ERP  & 76.05 & 78.62 & 52.01 & 51.36 & 40.92 & 0.41 \\
\textit{w/o.} IA-LoRA & 76.92 & 79.93 &  53.43 & 53.15 &  40.22 & 0.40  \\
\textbf{Crab(Ours)} & \textbf{78.94} & \textbf{80.15} & \textbf{59.00} & \textbf{54.44} & \textbf{41.78} & \textbf{0.42} \\ 

\bottomrule
\end{tabular}
}

\end{center}
\vspace{-2em}
\end{table*}

\end{document}